\documentclass{article}
\usepackage[nonatbib,final]{neurips_data_2023}
\usepackage[utf8]{inputenc}
\usepackage[T1]{fontenc}
\usepackage{amsfonts}
\usepackage{amsmath}
\usepackage{amssymb}
\usepackage{booktabs}
\usepackage{caption}
\usepackage{colortbl,array}
\usepackage{graphicx}
\usepackage{microtype}
\usepackage{multirow,bigdelim}
\usepackage{nicefrac}
\usepackage{fdsymbol}
\usepackage{verbatim}
\usepackage{titlesec}
\usepackage{url}
\usepackage[dvipsnames]{xcolor}
\usepackage{xspace}
\usepackage{enumitem}\setlist[itemize]{noitemsep, topsep=0pt}

\usepackage[breaklinks=true,bookmarks=false,colorlinks,backref]{hyperref}
\usepackage{cleveref}

\usepackage{pifont}%
\newcommand{\cmark}{\ding{51}}%
\newcommand{\xmark}{\ding{55}}%

\newcommand{\dataset}{EPIC~Fields\xspace}
\newcommand{\epic}{EPIC-KITCHENS\xspace}
\newcommand{\ie}{\emph{i.e.},\xspace}
\newcommand{\eg}{\emph{e.g.},\xspace}

\makeatletter
\renewcommand{\paragraph}{%
  \@startsection{paragraph}{4}%
  {\z@}{-0em}{-0.5em}%
  {\normalfont\normalsize\bfseries}%
}
\makeatother

\usepackage{etoc}
\usepackage{subcaption}

\titlespacing*{\section}{0pt}{0.5\baselineskip}{0.3\baselineskip}
\titlespacing*{\subsection}{0pt}{0.5\baselineskip}{0.3\baselineskip}
\titlespacing*{\subsubsection}{0pt}{0.5\baselineskip}{0.3\baselineskip}

\title{\dataset \\ Marrying 3D Geometry and Video Understanding}
\author{%
Vadim Tschernezki$^{\bigstar\varheartsuit\vardiamondsuit}$ $\quad$
Ahmad Darkhalil$^{\bigstar\clubsuit}$ $\quad$
Zhifan Zhu$^{\bigstar\clubsuit}$ \\
\textbf{David Fouhey}$^{\spadesuit}$ $\quad$
\textbf{Iro Laina}$^\varheartsuit$ $\quad$
\textbf{Diane Larlus}$^\vardiamondsuit$
$\quad$ \textbf{Dima Damen}$^\clubsuit$
$\quad$ \textbf{Andrea Vedaldi}$^\varheartsuit$ \\ \\
\noindent
$^{\varheartsuit}$VGG, University of Oxford $\quad^{\clubsuit}$University of Bristol\\ \noindent $^{\spadesuit}$New York University $\quad^{\vardiamondsuit}$NAVER LABS Europe  $\quad^{\bigstar}$: Equal Contribution}

\begin{document}
\maketitle

\begin{abstract}
Neural rendering is fuelling a unification of learning, 3D geometry and video understanding that has been waiting for more than two decades. Progress, however, is still hampered by a lack of suitable datasets and benchmarks. To address this gap, we introduce \dataset, an augmentation of \epic with 3D camera information. 
Like other datasets for neural rendering, \dataset removes the complex and expensive step of reconstructing cameras using photogrammetry, and allows researchers to focus on modelling problems.
We illustrate the challenge of photogrammetry in egocentric videos of dynamic actions and propose innovations to address them.
Compared to other neural rendering datasets, \dataset is better tailored to video understanding because it is paired with labelled action segments and the recent VISOR segment annotations.
To further motivate the community, we also evaluate three benchmark tasks in neural rendering and segmenting dynamic objects, with strong baselines that showcase what is not possible today. We also highlight the advantage of geometry in semi-supervised video object segmentations on the VISOR annotations. {}\dataset reconstructs 96\% of videos in EPIC-KITCHENS, registering 19M frames in 99 hours recorded in 45 kitchens, and is available from: \url{http://epic-kitchens.github.io/epic-fields}
\end{abstract}

\begin{figure}[h]
\centering
\includegraphics[width=\linewidth]{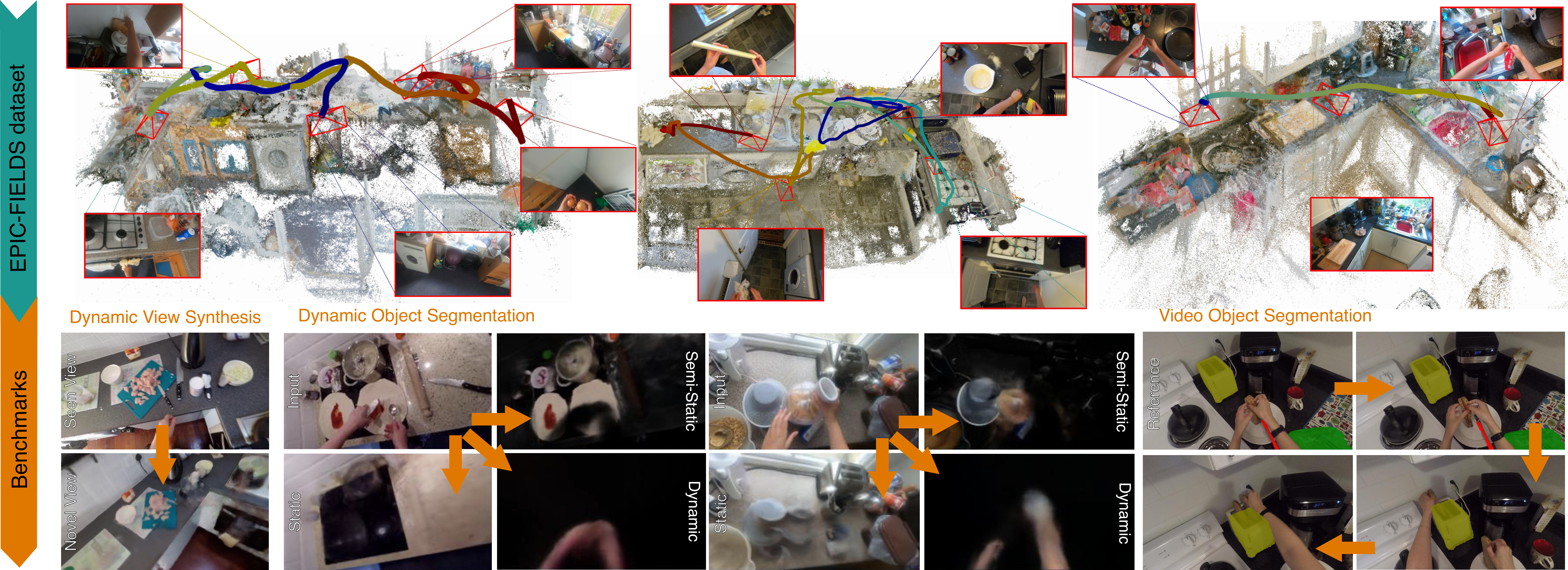}
\vspace{-1em}
\captionsetup{type=figure}
\captionof{figure}{
We propose \dataset that extends \epic with 3D information, including full frame-rate camera pose trajectories (top). These are directly obtained from dynamic sequences of object interactions (sampled frames) without additional modalities or pre-scans. We showcase \dataset through several benchmarks (bottom) that use the fusion of geometric and semantic cues.
}\label{fig:teaser}
\vspace*{-12pt}
\end{figure}

\section{Introduction}%
\label{s:introduction}

Recent breakthroughs in neural rendering~\cite{sitzmann19scene, mildenhall20nerf:} have enabled a deeper integration of machine learning in geometric tasks like 3D reconstruction and rendering, creating a new opportunity to bring 3D geometry and video understanding closer together.
By representing videos in 3D we can explain away the variability induced by the camera motion, which is dominant especially in egocentric videos.
We can also integrate information extracted from each frame independently into a global, consistent interpretation of the video, as demonstrated by semantic neural rendering~\cite{zhi21in-place, tschernezki22neural, kobayashi22decomposing, yang21learning, kundu22panoptic, fu22panoptic, vora21nesf:}.
However, such successes have been mostly limited to {\it static scenarios}, where only the camera moves.
Indeed, 3D reconstruction still struggles with dynamic content and much work remains before we can have a 3D understanding of dynamic phenomena like actions and activities.

An obstacle to further progress in 3D video understanding is the lack of suitable development data.
In this paper, we address this gap by introducing \emph{\dataset}, an extension of the popular \epic~\cite{damen18scaling} dataset which adds reconstructed 3D cameras and new benchmark tasks assessing both 3D reconstruction and semantic video understanding.

We choose to build on \epic because it is an established benchmark for 2D video understanding with rich annotations.
Furthermore, it contains egocentric videos which are likely to benefit from 3D understanding, but which also challenge existing 3D reconstruction techniques due to their highly dynamic content and long duration (up to one hour).
Dynamics include the motion of the actor and of the objects that they manipulate, as well as object transformations (\eg slicing a carrot).
Furthermore, most objects are mostly static, moving only during brief spells of active manipulation.
These challenges push the limits of current dynamic 3D reconstruction methods, which are usually restricted to short videos or focus on one class of objects~\cite{pumarola20d-nerf:,park21hypernerf:,park20deformable, li21neural, kwon21neural,song22nerfplayer:}.

Obtaining camera information for \epic is challenging since structure-from-motion methods often fail on such complex videos.
By solving this problem, we make it much easier for other researchers to start on 3D video understanding even when they are not experts in 3D vision, similar to what the setup and data introduced in works like NeRF~\cite{mildenhall20nerf:} has done for static 3D reconstruction.
Furthermore, while we propose specific benchmark tasks, we anticipate that researchers will be able to use our dataset to investigate many more questions than the ones we investigate here.

In summary, our first contribution is to augment \epic with camera information.
To overcome the limitations~\cite{Guan_2020_CVPR} of traditional structure-from-motion pipelines~\cite{schonberger16structure-from-motion}, which struggle with egocentric videos, we introduce a pre-processing step that intelligently sub-samples frames from these videos, resulting in higher reconstruction reliability and speed.
Our second contribution is to introduce new benchmark tasks that require or can benefit from the 3D cameras: dynamic novel view synthesis (\ie reconstructing unseen frames given a subset of frames from a monocular video); identifying and segmenting objects that move independently from the camera; and semi-supervised video object segmentation.
These benchmarks use and extend the VISOR~\cite{Darkhalil22VISOR} annotations to provide dense ground-truth semantic labels.
We report a number of baselines and conclude that, while 3D reconstruction can indeed benefit video understanding, existing approaches are challenged by the dynamic aspects of \dataset.

\section{Related work}%
\label{s:related}

\paragraph{Egocentric action understanding using 3D.}
Some egocentric datasets~\cite{Park_2016_CVPR,Damen2014,grauman2022ego4d} contain static 3D scans of the recording locations. These typically do not contain actions, or the environments are scanned post-hoc, usually with an additional step. For instance,~\cite{Park_2016_CVPR} uses stereo egocentric cameras, but no activities, and in~\cite{Damen2014,grauman2022ego4d}, reconstruction is done afterwards via hardware or additional dedicated scans. These scans are costly, which is why just 13\% of Ego4D~\cite{grauman2022ego4d} data comes with a 3D scan. In contrast, we provide a pipeline for estimating camera poses from egocentric data without additional hardware or scans, which we demonstrate on an existing, challenging dataset \epic.

\paragraph{Inferring cameras in egocentric videos.}

In this work, we perform the challenging task of reconstructing 3D camera poses from egocentric videos that show dynamic activities from a single camera. Since the \epic~\cite{Damen2022RESCALING} dataset is unscripted, the videos show natural interactions by participants in their homes, who act swiftly due to familiarity. Prior work~\cite{Guan_2020_CVPR,Nagarajan_2020_CVPR,tan2023egodistill} on these videos highlights the challenge. In~\cite{Guan_2020_CVPR}, where
ORB-SLAM was used to find short clips where the camera pose was stable, the authors note that bundle adjustment failed and reconstructions lasted for just 7 second intervals. Using~\cite{Guan_2020_CVPR},~\cite{Nagarajan_2020_CVPR} found hot-spots, but commented that just 44\% of the frames could be registered. Others have used additional hardware information; for instance, {}~\cite{tan2023egodistill} proposed using IMU data to establish short-term trajectories. In contrast, this work shows how to reconstruct cameras for \emph{full} videos in \epic, without additional assumptions, data, or hardware.

\paragraph{Multi-view videos.}
A different approach to enabling neural rendering is calibrated multiview setups. Many of these datasets, however, capture humans in a ``blank context'', including HumanEva~\cite{sigal10humaneva:},
Human3.6M~\cite{ionescu14human3.6m:},
AIST++~\cite{tsuchida19aist,li21learn}, and
ZJU-Mocap~\cite{peng20neural}. There are datasets capturing humans in complex environments, such as the Immersive Light Field dataset~\cite{broxton20immersive}, NVIDIA Dynamic Scene Datasets~\cite{yoon20novel}, UCSD Dynamic Scene Dataset~\cite{lin21deep}, and Plenoptic Video datasets~\cite{li22neural}. However, these videos are short (1--2 min) and, due to the capture setup, show actions outside of their natural environment.  In contrast, \epic is captured with an egocentric camera and shows long captures of indoor activities. Our contribution of reconstructing the cameras over time turns the egocentric data into the multiview data needed while retaining the naturalness of the data.

\paragraph{NeRF and dynamics.}  NeRF extensions to dynamic data can be roughly divided into
approaches that add time as an additional dimension of the radiance fields~\cite{martinbrualla2020nerfw,tschernezki21neuraldiff,xian21space-time,gao21dynamic,wang22fourier,li22neural,fridovich-keil23k-planes:,cao23hexplane:}
and those that instead model explicitly 3D flow and reduce the reconstruction to a canonical (static) one~\cite{pumarola20d-nerf:,park20deformable,yoon20novel,park2021nerfies,wang21neural,tretschk21non-rigid,li21neural,du21neural,yuan21star:,song22nerfplayer:,guo22neural,fang22fast,li22dynibar:,liu22devrf:}.
While these methods demonstrate successes, their success depends on the dominance of camera motion over scene motion~\cite{gao2022monocular}.
Scene motion by dynamic objects is not always common in existing datasets. Our proposed \dataset contains both camera motion and fast continuous motion by the actor visible in the camera's field of view.

\paragraph{NeRF and semantics.}

Authors have already noted that neural rendering and 3D geometry can be helpful allies of video understanding.
For instance,
Semantic NeRF~\cite{zhi21in-place,vora21nesf:} proposes to predict dense semantic labels in addition to RGB colours, while {}\cite{kundu22panoptic,fu22panoptic,siddiqui2023panoptic,wang2022dm} consider panoptic segmentations (things and stuff).
{}\cite{tschernezki22neural,kobayashi22decomposing,liang2023semantic} propose to fuse semantic features from pre-trained ViTs~\cite{caron21emerging,li2022languagedriven,touvron21training} into a neural reconstruction.
{}\cite{yang21learning,zhang2023nerflets} represent a scene as a composition of static objects given their 2D masks.
Several studies employ neural rendering to separate scenes into objects and background either without or with weak supervisory signals~\cite{fan2023nerfsos, xie2021fig, yu2022unsupervised, tschernezki21neuraldiff,sharma2023neural, ost2021neural, mirzaei2022laterf,wu2022d}.
With a few exceptions~\cite{tschernezki21neuraldiff, liang2023semantic, wu2022d}, however, little work has been done on decomposing \emph{dynamic} scenes into objects.

\section{The \dataset dataset}%
\label{s:method}

We introduce here the new \emph{\dataset} dataset.
We first describe the content of the dataset and then the process of constructing it, including several technical innovations that made it possible.

\subsection{\dataset in a nutshell}

\dataset extends \epic to include camera pose information.
\epic contains videos of cooking activities collected using a head-mounted camera in 45 different kitchens.
It has semantic annotations for fine-grained actions and their action-relevant objects, including 90K start-end times of actions~\cite{Damen2022RESCALING}. VISOR~\cite{Darkhalil22VISOR} adds 272K manually annotated masks and 9.9M interpolated masks of hands and active objects.
With \dataset, we further contribute camera extrinsic parameters for each video frame as well as camera intrinsic parameters.
Using the technique described in \Cref{subsec:dataset_construction}, we successfully processed 671 videos spanning all 45 kitchens, resulting in 18,790,333 registered video frames with estimated camera poses.

\begin{figure}
\centering
\includegraphics[width=\linewidth]{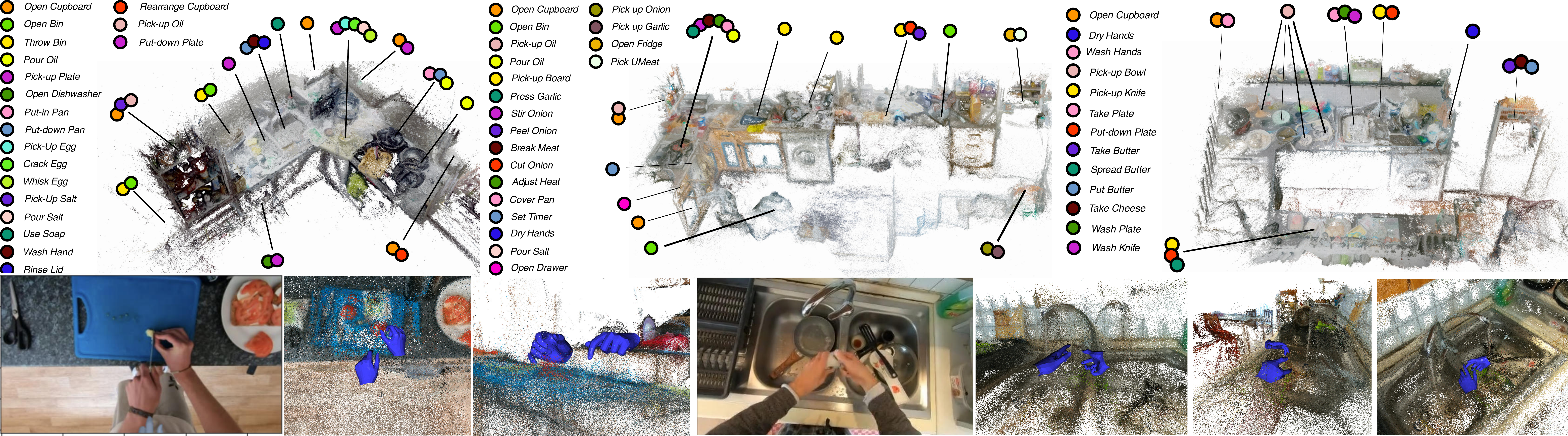}
\vspace{-1em}
\captionsetup{type=figure}
\captionof{figure}{
\dataset unlocks applications that combine interactions with 3D information. We showcase examples of actions grounded in 3D (top row), and examples of integrating single-image 3D hands~\cite{rong2020frankmocap} into the kitchen reconstruction during interactions (bottom row).}\label{fig:actions}
\end{figure}

\paragraph{Motivation.}
Our camera annotations facilitate reconstructing and interpreting videos in 3D.
\Cref{fig:actions} illustrates this point by mapping some 2D action annotations from \epic to the 3D space.
Lifting annotations to 3D puts them in the wider context of the environment where actions occur,
and enables studying the relevance of 3D egocentric trajectories to actions (for anticipation), objects (for understanding object state changes), and hand-object understanding.
The figure also illustrates mapping hand meshes extracted using~\cite{rong2020frankmocap} to the 3D context of the kitchen.

\paragraph{Ethics, licensing, data protection.}

\epic was collected with ethics approval by the University of Bristol and explicit consent from the participants.
The data does not contain personal identifiable information or offensive content and is provided under a non-commercial license.
\dataset is released under the same terms.

\subsection{Dataset construction}%
\label{subsec:dataset_construction}

Because \epic videos were not collected with 3D reconstruction in mind they are difficult to reconstruct.
For instance, they contain many dynamic objects:
hands are visible in 95\% of the frames and the focus of attention is often an object actively manipulated.
Standard reconstruction pipelines operate under the assumption that the scene is static and are thus only moderately robust to dynamic objects.
Other challenges include the video length ($\sim$9~mins on average) and the skewed distribution of viewpoints: videos %
alternate phases of small motion around hot-spots (\eg cooking at a hob or washing at the sink) and fast motion between hot-spots (\eg moving the pot to the sink).

We address these challenges by:
(1) filtering videos to reduce the number of redundant frames, computational cost, and skew;
(2) using structure from motion~(SfM) to reconstruct the scene from the filtered frames;
(3) registering the remaining frames to the sparse reconstruction.
We accept a video's reconstruction if $70\%$ or more of its frames are registered successfully.
In this manner, we can reconstruct 96\% of all \epic videos.
We next describe each step, with details in the supplement.

\newcommand{\HB}{\mathcal{HB}}%

\paragraph{Frame filtering.}

The goal of frame filtering is to downsample a video to reduce redundancy and skew while maintaining sufficient viewpoint coverage for accurate reconstruction.
We filter frames by seeking temporal windows where frames have substantial visual overlap and then only keep one frame per window, similar to redundant frame mining~\cite{schonberger16structure-from-motion,tang2017gslam} and other SfM or SLAM pipelines.
Overlap between frames is measured by estimating homographies by matching SIFT features~\cite{d.lowe04distinctive}.
Given a homography $H$ between two frames, we define their visual overlap $\tilde{r}$ to be the fraction of the first frame area covered by the quadrilateral formed by warping the second frame corners by $H$.
Windows are formed greedily, finding runs of frames $(i+1, \ldots, i+k)$ with overlap $\tilde{r} \ge 0.9$ to the first frame $i$ and discarding them.
Filtering discards on average about $82\%$ of frames in each video while also retaining a sufficient number of frames in the critical transitions between hot-spots.

\paragraph{Sparse reconstruction.}

The filtered frames are fed to an off-the-shelf structure-from-motion pipeline.
Among these, we found COLMAP~\cite{schonberger16structure-from-motion} to be more effective than VINS-MONO~\cite{QinVINSMONO}, which suffered from frequent drifts and restarts.

\begin{figure*}[t]
\centering
\includegraphics[width=1.0\textwidth]{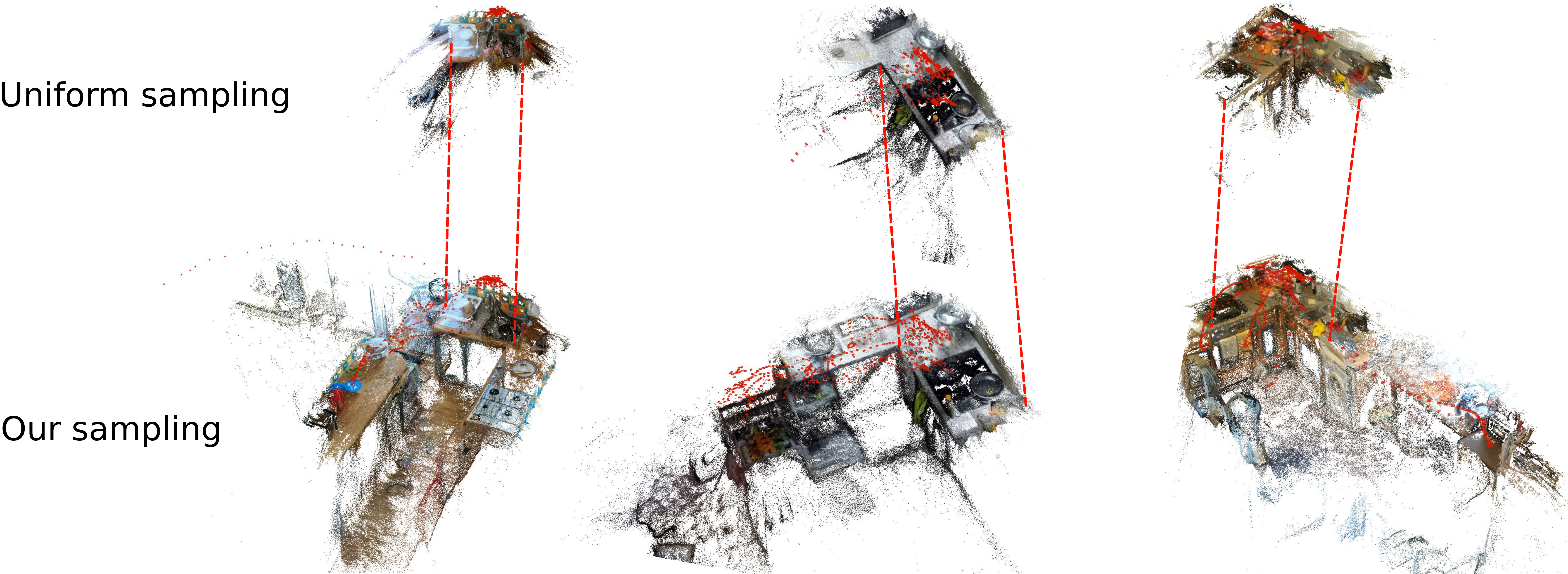}
\caption{
\textbf{3D reconstructions with different sampling.}
We compare three scenes reconstructed using either uniform frame selection or our homography-based pipeline.
Uniform sampling yields partial reconstructions with limited coverage. 
Ours demonstrates superior performance, resulting in better coverage by registering successfully more viewpoints. 
}\label{fig:qual_sampling}
\vspace*{-12pt}
\end{figure*}

In \Cref{tab:eval_filtering} we analyse the effectiveness of the homography-based filtering algorithm by comparing it to a na{\"\i}ve filter that subsamples frames uniformly.
We use 30 randomly selected videos for this experiment and report two standard SfM metrics~\cite{schonberger16structure-from-motion}: the average reprojection error and the number of 3D points in the reconstruction.
The first metric is a proxy for the accuracy of the reconstruction, and the second for its coverage. 
Both filtering techniques reduce the number of frames equally and thus result in similar computational complexity.
However, homography-based filtering also addresses the skew and results in a significantly better success rate, increased coverage, and reduced reprojection error compared to uniform subsampling.
Besides considering the number of points reconstructed, \Cref{fig:qual_sampling} shows qualitatively the notably improved coverage obtained by homography-based filtering.

\begin{table}[t]
\small
\centering
\caption{\textbf{Impact of frame filtering on the reconstruction quality.}
We compare the sparse reconstruction of 30 videos using either homography-based or uniform frame filtering.
Na{\"\i}ve uniform sampling
results in only 27 of the 30 videos being reconstructed successfully (\ie dense registration rate $\ge70\%$).
Furthermore, the successful reconstructions have significantly reduced coverage (-16.64\%) and increased reprojection error (+4.76\%) compared to homography-based filtering.
}
\vspace{0.5em}
\begin{tabular}{@{}lc@{\hskip 10pt}c@{\hskip 10pt}c@{\hskip 10pt}c@{\hskip 10pt}c@{}}
\toprule
\multirow{2}{*}{\textbf{Frame sampling}} & {\textbf{Avg. \#3D}} & \textbf{Avg. Repr.} &\textbf{Avg. Reg.} & \textbf{Successful} \\
 & \textbf{Points} & \textbf{Error} & \textbf{Rate}& \textbf{Reconstructions} \\
\midrule
Homography-based (ours) & 27,763 & 0.798 & 98.6\%& 30/30 \\
Uniformly &23,142 & 0.836 & 89.0\%&27/30 \\
\midrule
Relative change  & -16.64\%  & 4.76\%  & 9.77\% & -10 \% \\
\bottomrule
\end{tabular}\label{tab:eval_filtering}
\end{table}

\paragraph{Dense reconstruction, automated verification, and restart.}

After obtaining the sparse reconstruction from the filtered subset of video frames, we use COLMAP to register the remaining frames against it, which is computationally cheap.
We accept the final reconstruction if $\geq$70\% of the video's frames, at full frame rate, are registered successfully.
This process succeeds in 90\% (631 videos) of cases.
When a video is rejected, the reconstruction process is attempted again with a higher threshold $\tilde{r} \ge 0.95$; this usually doubles the number of frames that COLMAP needs to process for the reconstruction, but increases the success rate to 96\%.
We discuss reasons for the failure of the last 29 \epic videos in the supplement.

\paragraph{Application to other egocentric videos.}

While we developed our reconstruction pipeline by considering the \epic data, the approach we obtained is general and applies equally well to other egocentric video collections such as Ego4D~\cite{grauman21ego4d}, at least for indoor locations.
We give examples of these reconstructions in the supplement.

\section{The \dataset benchmarks, experiments and results}%
\label{s:bench}

We define three benchmarks on \dataset that probe 3D video understanding.
Annotations, evaluation code and baselines are released as part of \dataset; further details are in the supplement.

\subsection{Dynamic New-View Synthesis (D-NVS)}%
\label{s:new-view}

Given a subset of video frames as input, the goal of dynamic new-view synthesis (D-NVS) is to predict other video frames given only their timestamps and camera parameters.
While other D-NVS benchmarks exist, \dataset is more challenging due to the first-person perspective and the large number of dynamic objects.
In \Cref{tab:datasets} we compare \dataset to commonly used datasets in D-NVS\@.
\dataset offers a significant step up in complexity and scale with \textit{significantly longer} videos and associated semantics.
For detailed statistics, please refer to the supplement.

\begin{table*}[t!]
\caption{Comparison of datasets commonly used in dynamic new-view synthesis.\label{tab:datasets}
}\label{t:dataset_comparison}
\centering
\small
\begin{tabular}{lccccc}
\toprule
\textbf{Dataset}     & \textbf{\#Scenes}       & \textbf{Seq.~Length}   & \textbf{Monocular}         & \textbf{Semantics}   \\
\midrule
Nerfies~\cite{park2021nerfies}      & 4         & 8--15 sec            &   \xmark             &    \xmark   \\
D-NeRF~\cite{pumarola20d-nerf:}     & 8         & 1--3 sec             &   \xmark             &    \xmark   \\
Plenoptic Video~\cite{li22neural} & 6         & 10--60 sec            &   \xmark             &    \xmark    \\
NVIDIA Dynamic Scene Dataset~\cite{yoon20novel}      & 12         & 1--5 sec             &    4 / 12              &    \xmark   \\
HyperNeRF~\cite{park21hypernerf:}   & 16        & 8--15 sec            & 13 / 16           &    \xmark   \\
iPhone~\cite{gao2022monocular}      & 14        & 8--15 sec            & 7 / 14            &    \xmark   \\
SAFF~\cite{liang2023semantic} & 8 & 1--5sec & \xmark & \cmark \\
\midrule
\multirow{1}{*}{\textbf{\dataset} \textbf{[D-NVS]} (ours)}         & \multirow{1}{*}{50}        & 6--37 min (Avg 22)      &    \multirow{1}{*}{50 / 50}        &  \multirow{1}{*}{\cmark}    \\
\bottomrule
\end{tabular}
\end{table*}

\definecolor{easy}{RGB}{112,173,71}
\definecolor{med}{RGB}{241,183,119}
\definecolor{hard}{RGB}{192,0,0}

\begin{figure*}[t]
\centering
\includegraphics[width=0.99\textwidth,trim={0 3.7cm 2.2cm 0},clip]{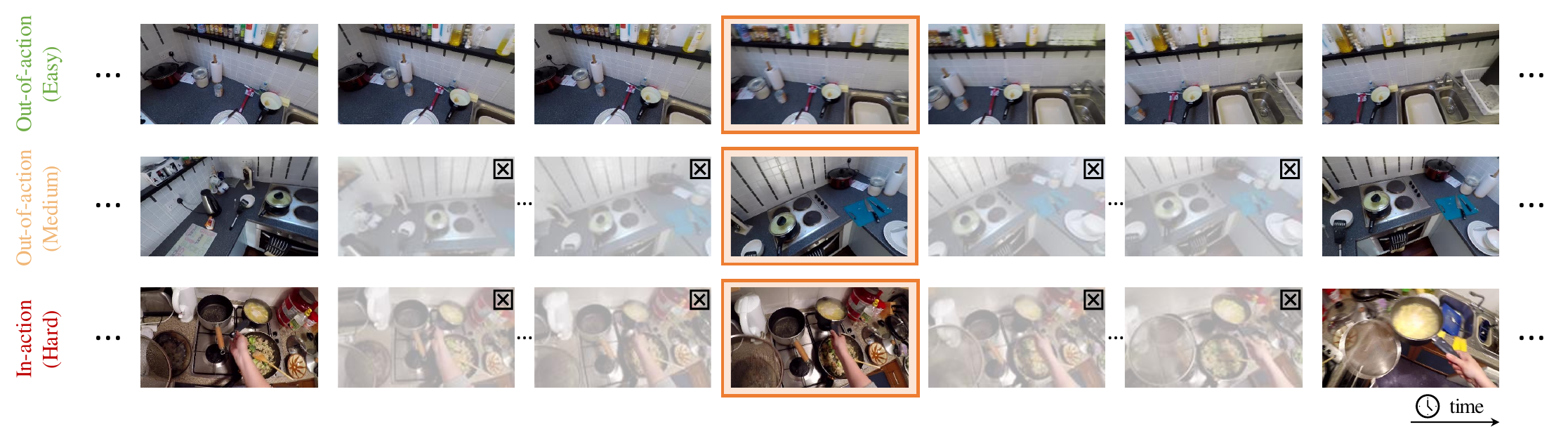}
\vspace{-0.5em}
\caption{ 
\textbf{Definition of the three difficulty levels for the task of dynamic new-view synthesis.}
Validation and test frames are selected to meet three reconstruction difficulty levels.
\textcolor{hard}{\bf In-Action frames (Hard)} happen during an action and are harder to reconstruct due to the dynamics.
\textcolor{med}{\bf Out-of-Action (Medium) frames} happen outside an action, but are far from a train frame.
\textcolor{easy}{\bf Out-of-Action (Easy) frames} are near train frames.
Frames in a bounding box (orange) represent either val/test frames.
Frames marked with a cross are discarded to create a larger time gap around each val/test frame (medium and hard levels).
All other frames can be used for training.}%
\label{fig:difficulty}
\vspace*{-12pt}
\end{figure*}

\paragraph{Video selection.}

Due to the computational cost of most D-NVS algorithms, we limit the D-NVS benchmark to a subset of 50 videos (14.7 hours and 2.86M registered frames) extracted from the train/val set of VISOR~\cite{Darkhalil22VISOR} (this selection includes 96.1\% of the frames annotated in VISOR).

\paragraph{Frame selection.}

For each video in the D-NVS benchmark, we select the video frames to be used as input to the system (training) and those that remain unseen and are used for evaluation only (validation/testing).
Specifically, we propose categorising evaluation frames into three tiers of difficulty (easy, medium, hard\,---\,visualised in \Cref{fig:difficulty}), determined by the type of motion and the temporal gap between the evaluation and training frames.
\textbf{In-Action} frames correspond to common `put', `take', and `cut' actions annotated in \epic, based on their start-stop times; they are characterised by substantial object motion due to hand-object interactions and are thus more difficult to reconstruct.
In pursuit of a greater challenge, for the \textbf{In-Action (Hard)} set of frames, we exclude frames from the training set occurring within 1 second of a test frame.
\textbf{Out-of-Action} frames occur outside action segments, where there is no appreciable motion except for the camera, making these frames generally easier to reconstruct.
For the \textbf{Out-of-Action (Medium)} set, we sample 70\% of the out-of-action frames with the same time gap as above.
The \textbf{Out-of-Action (Easy)} set corresponds to the remaining 30\% without removing the neighbouring training frames.
The reasoning is that it is generally easier to predict a frame temporarily close to a training one.
We assign every other evaluation frame to the validation and test sets, respectively.
The average time gap between consecutive evaluation frames is 3.73 seconds.
Further statistics are provided in the supplement.

\paragraph{Benchmark methods.}

To demonstrate how \dataset can be used and to probe the limits of 
the state of the art in such challenging scenarios, we consider three neural rendering approaches:
NeRF-W~\cite{martinbrualla2020nerfw}, NeuralDiff~\cite{tschernezki21neuraldiff}, and T-NeRF+, an extended version of T-NeRF~\cite{gao2022monocular}.

\emph{NeuralDiff}~\cite{tschernezki21neuraldiff} is a method tailored to egocentric videos.
It uses three parallel streams to separate the scene into the actor, the transient objects (that move at some point in the video), and the background that remains static.
We combine the predictions of the actor and transient objects to predict our \emph{dynamic} and \emph{semi-static} objects, which will be relevant in \Cref{subsec:udos}.

\emph{NeRF-W}~\cite{martinbrualla2020nerfw} 
augments NeRF with the ability to `explain' photometric and environmental (non-constant) variations by learning a low-dimensional latent space that can modulate scene appearance and geometry.
As a result, NeRF-W also separates static and transient components.
We follow the modification from~\cite{tschernezki21neuraldiff} to render NeRF-W applicable to video frames and the D-NVS task.

\emph{T-NeRF+}~\cite{gao2022monocular} was proposed as a baseline to evaluate state-of-the-art NeRFs on dynamic scenes.
It was shown to outperform other methods
in terms of the quality %
of the synthesised images.
We extend T-NeRF by adding another stream to the time-conditioned NeRF architecture that models the background (static parts of the scene).

\begin{figure}[t]
\centering
\includegraphics[width=\textwidth]{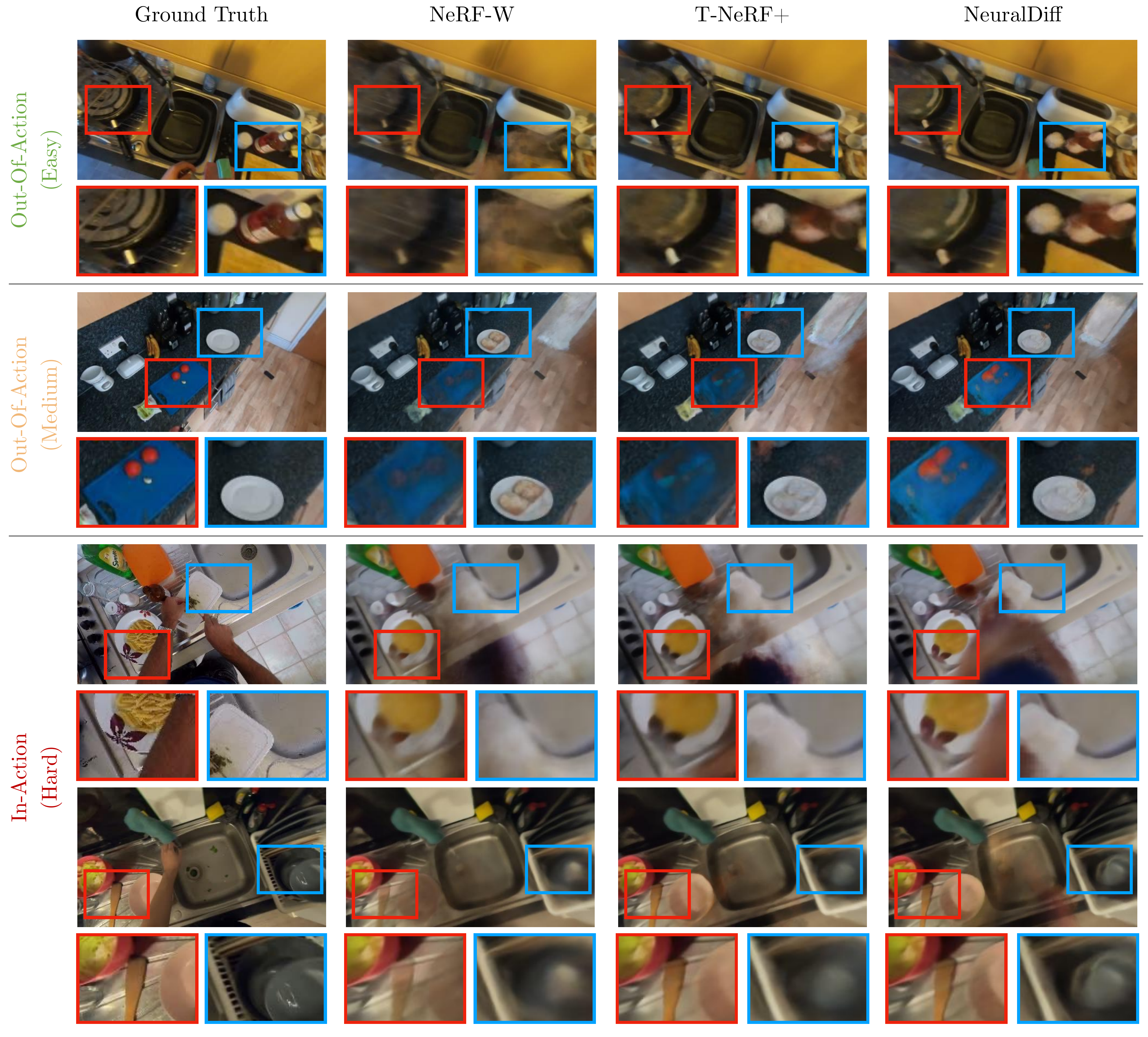}
\caption{\textbf{Dynamic new-view synthesis.}
We compare the outputs of 3D methods NeRF-W~\cite{martinbrualla2020nerfw}, T-NeRF+~\cite{gao2022monocular}, and NeuralDiff~\cite{tschernezki22neural}, for novel viewpoints, across three different complexity levels.
The predictions are more accurate with less difficult motion as shown in the first and second row.
The task becomes more challenging for our hard samples.}\label{fig:qual_dvs}
\vspace*{-12pt}
\end{figure}

\paragraph{Results.}
To measure performance on this task, we report the Peak Signal-to-Noise Ratio (PSNR) of the test frame reconstructions, which is a proxy for the quality of the underlying 3D reconstructions with the key advantage of not requiring 3D ground-truth for evaluation.
We report results in \Cref{tab:quant_dvs} for the three levels of difficulty.
There is a strong relationship between PSNR and difficulty: PSNR is consistently lower for all methods when rendering views during actions (hard) compared to outside actions (medium, easy). 
Some limitations of rendering these hard test frames are shown in \Cref{fig:qual_dvs}.
For example, the bottom row shows that no 3D baseline renders the person's arm correctly, since all models struggle to interpolate the person's movement between frames.
We further observe a significant gap in rendering quality if we calculate PSNR separately for foreground and background regions.
We use the VISOR annotations of hands and active objects for In-Action frames to obtain this separation. 
These results not only highlight the existing limitations of current methods but also offer a valuable benchmark for assessing potential improvements in a targeted manner.

\subsection{Unsupervised Dynamic Object Segmentation (UDOS)}%
\label{subsec:udos}

The goal of Unsupervised Dynamic Object Segmentation (UDOS) is to identify which regions in each frame correspond to dynamic objects.
This task can be approached in 2D only but is a good proxy to assess 3D methods as well, and can, in fact, be boosted by 3D modelling.
Here, we extend the setting introduced in~\cite{tschernezki21neuraldiff}, using 20$\times$ more data and adopting a more nuanced evaluation protocol.

\begin{table}[t]
\caption{\textbf{Dynamic new-view synthesis}. We compare different neural rendering approaches for frames from different difficulty levels (easy, medium, hard). We report PSNR considering all pixels in each test frame. Given the mask annotations from VISOR for \emph{In-Action} frames, we also report PSNR on background (BG) and foreground (FG) pixels separately for the hard (\emph{In-Action}) setting. 
}\label{tab:quant_dvs}
\small
\centering
\setlength{\tabcolsep}{8pt}
\begin{tabular}{l@{\hskip 30pt}ccccc} %
\toprule
\multirow{2}{*}{\textbf{Method}} & \multirow{2}{*}{\textbf{Easy}} & \multirow{2}{*}{\textbf{Medium}} & \multicolumn{3}{c}{\textbf{Hard}} \\ 
\cmidrule(lr){4-6}
& & & All & BG & FG \\
\midrule
NeRF-W~{\protect\NoHyper\cite{martinbrualla2020nerfw}\protect\endNoHyper}        & 21.13     & 19.3    & 17.93    & 18.99 & 13.54             \\
T-NeRF+~{\protect\NoHyper\cite{gao2022monocular}\protect\endNoHyper}             & 21.58     & 19.81    & 18.44    & 19.73  & 13.74            \\

NeuralDiff~{\protect\NoHyper\cite{tschernezki21neuraldiff}\protect\endNoHyper}   & 22.14     & 19.88    & 18.36    & 19.54   &  13.37            \\ 
\bottomrule
\end{tabular}
\vspace{-12pt}
\end{table}

\paragraph{Video and frame selection.}

We use the same selection of videos as for the D-NVS task, but only use the In-Action frames with VISOR annotations, as they provide ground-truth dynamic object segmentations. 
We convert VISOR masks into a foreground-background mask for each frame in three ways, depending on objects that are currently moving, or those that have moved at a different time in the video.
In the \textbf{dynamic objects only} setting, the foreground contains hands and other visible body parts as well as object masks only for objects that are currently being moved.
We use the VISOR contact annotations to identify these objects and augment these with additional manual masks for visible body parts including torsos, legs, and feet.
More details are in the supplement.
In the \textbf{semi-static only} setting, we consider only objects that moved at some point during the video, but not during the current frame.
We select these objects by watching the video and identifying all objects that have moved at least once.
VISOR contains annotations of these objects only on frames where they are considered \emph{active}.
We employ an automated method to propagate the annotations to cover all frames, resulting in a set of semi-static object masks.
This is the %
complete set of masks for all objects that have moved at any point in the video, even if they are temporarily static.
More details can be found in the supplement.
We combine both to report the 
\textbf{dynamic and semi-static} setting.

\paragraph{Benchmark methods.}

We use NeuralDiff and NeRF-W from the NVS task, since, by design, they decompose scenes into static and dynamic components.
Additional considerations are necessary to make \emph{T-NeRF+} applicable to UDOS\@.
In order to disentangle the modelling of both radiance fields in terms of temporal variation, we apply the uncertainty modelling from~\cite{martinbrualla2020nerfw} to model a change in observed colours of pixels that occur due to dynamic effects inside the scene.
This extension enables \emph{T-NeRF+} to learn a decomposed radiance field.

We also consider a 2D baseline, \emph{Motion Grouping (MG)}~\cite{yang2021self}, a state-of-the-art method for self-supervised video object segmentation.
It trains a segmentation model using an autoencoder-like framework.
The model has two output layers; one layer represents the background, and the other layer identifies one or more moving objects in the foreground, including their opacity layers.
These layers are then linearly composed and optimised to reconstruct the input image.
Since this approach is unsupervised, it can be compared fairly to the 3D baselines for this task.

\paragraph{Results.}

To evaluate performance, we measure 2D segmentation accuracy on test frames using mean average precision (mAP) as in~\cite{tschernezki21neuraldiff}.
\Cref{tab:quant_seg} compares unsupervised 2D baselines and 3D baselines.
Depending on the type of observed motion, 3D-based methods offer advantages over 2D methods and vice versa.
For example, 3D-based methods are better suited for discovering semi-static objects that are not currently in motion, \ie they have been moved at different times within a video. 
This is evident by the improved segmentation performance when considering this type of motion (\ie \textit{SS+D} and purely SS).
However, we note that none of the 3D-based methods explicitly consider motion. 
Consequently, MG, which takes as input optical flow, performs better on purely dynamic motion, but struggles to segment objects that are temporarily not moving.
This drawback of 3D-based methods, compared to 2D motion-based methods, underscores the current challenge in capturing dynamics in neural rendering. Addressing this limitation is an open question for future research.

\begin{figure*}[t]
\centering
\includegraphics[width=\textwidth]{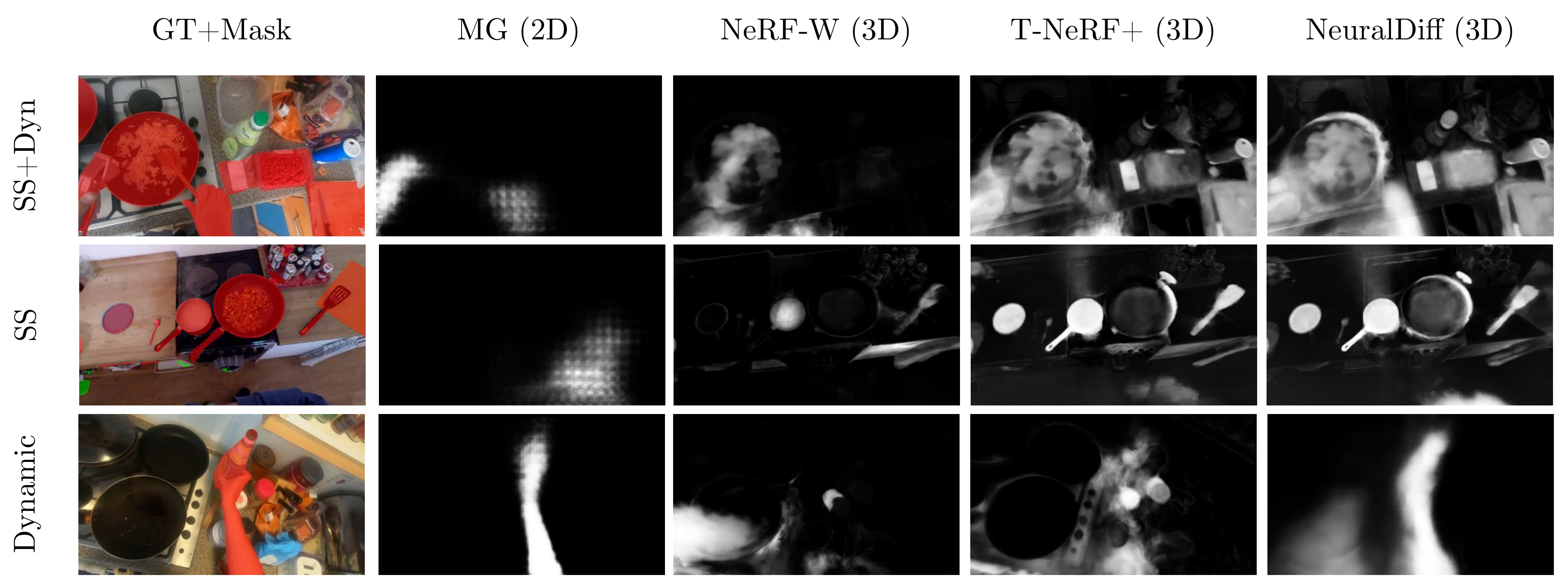}
\caption{\textbf{Unsupervised dynamic object segmentation.}
We compare three 3D baselines (NeRF-W~\cite{martinbrualla2020nerfw}, T-NeRF+~\cite{gao2022monocular} and NeuralDiff~\cite{tschernezki22neural}) and one 2D baseline (MG~\cite{yang2021self}) with three motion types. The 2D baseline captures the person and does well on the purely dynamic (short-range motion) setting.
3D models do this and also segment semi-static (SS) components (long-range motion, i.e., objects that were moved some time ago). In \textit{SS+Dyn}, the evaluation includes SS and dynamic components.}\label{fig:qual_segment}
\end{figure*}

\Cref{fig:qual_segment} shows qualitative results.
We observe that MG performs particularly well on objects that are constantly in motion, for example, the moving body parts of the person.
Among the 3D methods, NeuralDiff is better at capturing dynamic objects, and, unlike MG, both NeuralDiff and T-NeRF+ are able to segment various semi-static objects as well since they do not rely on continuous motion.

\subsection{Semi-Supervised Video Object Segmentation (VOS)}

Semi-Supervised Video Object Segmentation (VOS) is a standard semi-supervised video understanding task:
given the mask for one or more objects in a reference frame, the goal is to propagate the segments to subsequent frames.
For this task, we use the train/val splits published as part of the VISOR VOS benchmark (See~\cite{Darkhalil22VISOR} Sec. 5.1).
VOS is usually approached by using 2D models.
Here, we explore how the 3D information in \dataset can be used for it instead.

\paragraph{Benchmark methods.} 
We evaluate two na{\"\i}ve baselines for this task, one in 2D and another in 3D. For completeness, we also compare these to existing, trained 2D VOS models.

\newcommand{\imprv}[1]{{\small{\color{ForestGreen}{(#1)}}}}
\newcommand{\decr}[1]{{\small{\color{Red}{(-#1)}}}}

\begin{table*}[t]
\caption{\textbf{Unsupervised dynamic object segmentation}. We report the mean average precision~(mAP) on segmenting the semi-static (SS) and dynamic components of the scene, and also their union (SS+Dyn). All methods are trained without explicit supervision, \ie no masks are used during training,
only for evaluation.}\label{tab:quant_seg}
\small
\centering
\setlength{\tabcolsep}{6pt}
\begin{tabular}{lc@{\hskip 30pt}ccc} %
\toprule
\textbf{Method} & \textbf{3D} & \textbf{SS+Dyn} & \textbf{SS} & \textbf{Dynamic}\\ 
\midrule
MG~{\protect\NoHyper\cite{yang2021self}\protect\endNoHyper}     & \xmark & 55.53 & 12.78 & 64.27 \\
NeRF-W~{\protect\NoHyper\cite{martinbrualla2020nerfw}\protect\endNoHyper}     & \cmark  & 45.62  & 20.97 & 28.52             \\
T-NeRF+~{\protect\NoHyper\cite{gao2022monocular}\protect\endNoHyper}        &   \cmark  & 64.91  & 24.48 & 44.27 \\
NeuralDiff~{\protect\NoHyper\cite{tschernezki21neuraldiff}\protect\endNoHyper}  & \cmark & 69.74 & 25.55 & 55.58 \\ 
\bottomrule
\end{tabular}
\end{table*}

\emph{Fixed in 2D.}
We make the assumption that the pixels in the first frame remain constant throughout the entire sequence.
This na{\"\i}ve baseline is prone to failure when the camera undergoes movement. %

\emph{Fixed in 3D.} %
To better understand the potential of 3D information for VOS, we compare the 2D baseline above to a 3D one.
In the 3D baseline, an object mask is projected to 3D and its position in 3D is fixed throughout the sequence. 
The mask is then re-projected to other frames using the available camera information.
This works well for static objects and achieves two effects.
First, objects can be reliably tracked over occlusions.
Second, detecting when these objects are in or out of view is a by-product of estimated camera poses.

\emph{Trained 2D models.} We also evaluate two state-of-the-art models for video object segmentation, STM~\cite{oh2019video} and XMEM~\cite{cheng2022xmem}.
These are trained on the train set of VISOR\@.

\begin{table*}[t]
\caption{\textbf{Semi-Supervised VOS}. We compare naive baselines in 2D and 3D, as well as pretrained/fine-tuned models on static and dynamic objects on the validation set of VISOR VOS\@.
*: two videos from the validation set are excluded as they don't have successful reconstructions.}\label{tab:task_vos}
\centering
\small
\resizebox{\linewidth}{!}{
\begin{tabular}{lc@{\hskip 1cm}llllllllll}
\toprule
\multirow{2}{*}{\textbf{Method}} & \multirow{2}{*}{\textbf{3D}} &
&
\multicolumn{3}{c}{\textbf{VISOR VAL\protect\NoHyper\cite{Darkhalil22VISOR}\protect\endNoHyper}} & \multicolumn{3}{c}{\textbf{Static}} & \multicolumn{3}{c}{\textbf{SS + Dyn}} \\
\cmidrule(lr){4-6} \cmidrule(lr){7-9} \cmidrule(lr){10-12}
 & & & $\mathcal{J}$\&$\mathcal{F} $ & $\mathcal{J}$  & $\mathcal{F}$  & $\mathcal{J}$\&$\mathcal{F} $ &$\mathcal{J}$  & $\mathcal{F}$ & $\mathcal{J}$\&$\mathcal{F} $&$\mathcal{J}$  & $\mathcal{F}$ \\
\midrule

Fixed in 2D
& \xmark & & 12.5 & 13.4  & 11.6  &
17.8 & 23.8 & 11.6 &
12.0 & 11.9 & 12.0  \\

Fixed in 3D *
& \cmark & & 31.3 & 30.5  & 32.2
& 48.4 & 52.2 & 44.6 &
29.6 & 27.8 & 31.5  \\

\midrule
Pretrained STM
& \xmark & & 63.0 & 60.8  & 65.2  &
64.3 & 65.4 & 63.1 &
63.7 & 60.8 & 65.5  \\

Fine-tuned STM
& \xmark & & 76.4 & 74.2  & 78.6
& 76.8 & 77.7 & 76.0
& 76.6 & 73.8 & 79.5  \\

Pretrained XMEM
& \xmark & & 64.0 & 61.5  & 66.4  &
63.2 & 64.0 & 62.5 &
64.1 & 61.1 & 67.1  \\

Fine-tuned XMEM
& \xmark & & 77.3 & 75.2  & 79.4
& 77.0 & 77.7 & 77.4
& 78.0 & 75.3 & 80.7  \\

\bottomrule
\end{tabular}
}
\end{table*}

\paragraph{Results.}

\begin{figure*}[t]
\centering
\includegraphics[width=\textwidth]{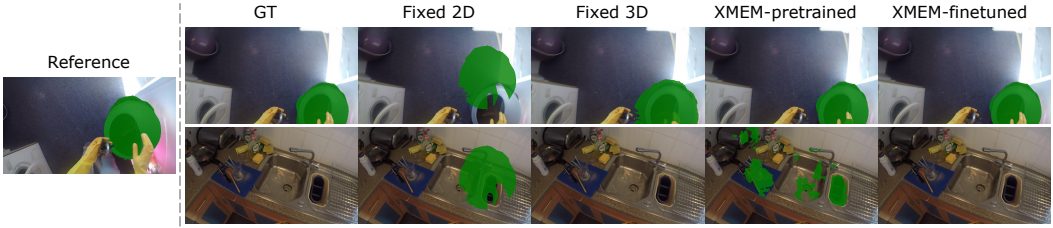}
\caption{\textbf{Semi-Supervised Video Object Segmentation.}
We compare our baselines on two frames from the same sequence.
The \textit{Fixed in 3D} baseline can track the bin over camera motion and recognise in/out-of view. Pretrained models usually suffer from false positives in the out-of-view scenes.}\label{fig:vos_qualitative}
\vspace*{-12pt}
\end{figure*}

We compare the baselines on the VISOR benchmark using the evaluation metrics defined in~\cite{pont20172017} which are the region similarity $\mathcal{J} $  and contour accuracy $\mathcal{F} $. 
We also distinguish the set of objects that are static, such as `fridge', `floor', and `sink', and report the above metrics separately for these and all other movable objects (SS+Dyn). %
\Cref{tab:task_vos} shows the results where the \textit{Fixed in 3D} clearly outperforms the \textit{Fixed in 2D} by a significant margin for the anticipated \textit{static} objects (+30.6\%) but also improves results for the remaining \textit{semi-static and dynamic} (+17.6\%) objects. This is because such objects do remain unmoved for some duration of the videos. This highlights the additional value derived from representing objects in 3D. \Cref{fig:vos_qualitative} visualises one example where the bin is successfully propagated using the \textit{Fixed in 3D} baseline, including when out of view. The pretrained models struggle to propagate masks for the novel objects in the dataset or for masks that go out of the scene. These are cases that the \textit{Fixed in 3D} baseline successfully handles. However, the fine-tuned models are quantitatively and qualitatively superior as they are trained on the dataset.
No prior work has utilised 3D information along with learnt models for the task of semi-supervised VOS\@.
We hope our novel benchmark can trigger new VOS approaches that tackle the combined challenge of keeping track of static objects in 3D and dynamic objects through trained propagation of objects during motion and transformations.

\section{Conclusions}\label{s:conclusions}

We introduced \dataset, a dataset to study 3D video understanding.
We addressed the difficult challenge of reconstructing cameras in \epic videos, introducing filtering and other techniques that are portable to other similar reconstruction scenarios.
Using these pre-computed cameras facilitates working on 3D video understanding even without significant expertise in photogrammetry.

With \dataset we also defined three benchmark tasks: dynamic new-view synthesis, unsupervised dynamic object segmentation, and video object segmentation.
Our results show that the performance of state-of-the-art dynamic neural reconstruction/rendering methods strongly depends on the type of motion.
In particular, the gap in reconstruction quality between the dynamic and the static parts of the videos show that there is ample margin for further improvements in the handling of dynamic objects.
Similar findings apply to the segmentation of objects, where 3D-based models can assist unsupervised video object segmentation and propagate masks of static objects over time.
We hope that these results, the proposed benchmark data and code (comprising evaluation, camera reconstruction, and baselines) will assist the community in investigating further methods that combine geometry and video understanding.

\paragraph{Societal impact.}

While we expect that our benchmark will lead to positive impact, including applications to augmented and mixed reality including AR assistants, there are potential negative impacts as well: better AR may be used for deception and many capabilities powering an assistant may also aid surveillance.

\newpage
\paragraph{Acknowledgements}
Project supported by EPSRC Program Grant Visual AI EP/T028572/1.
A.~Darkhalil is supported by EPSRC DTP program.
Z.~Zhu is supported by UoB-CSC Scholarship.
I.~Laina and A.~Vedaldi are supported by ERC-CoG UNION 101001212.
D.~Damen is supported by EPSRC Fellowship UMPIRE~EP/T004991/1.
D.~Fouhey is supported by a NSF CAREER~\#2142529.

We thank Sanja Fidler and Amlan Kar, from the University of Toronto, for contributions to the initial idea and discussions of this project.
We also thank Rhodri Guerrier for assisting in manual annotations of the test set.

\bibliographystyle{ieee_fullname}\bibliography{vedaldi_specific,vedaldi_general_cloned,more}

\appendix
\section*{Checklist}

\begin{enumerate}

\item For all authors\dots
\begin{enumerate}
  \item Do the main claims made in the abstract and introduction accurately reflect the paper's contributions and scope?
    \answerYes{We believe that our abstract and introduction accurately describe the work.}
  \item Did you describe the limitations of your work?
    \answerYes{We have discussed the limitations in the experiments. Indeed, one of the points of our paper is to show current limitations in existing work.}
  \item Did you discuss any potential negative societal impacts of your work?
    \answerYes{We have discussed potential negative impacts in the conclusions of the paper.}
  \item Have you read the ethics review guidelines and ensured that your paper conforms to them?
    \answerYes{}
\end{enumerate}

\item If you are including theoretical results\dots
\begin{enumerate}
  \item Did you state the full set of assumptions of all theoretical results?
    \answerNA{There are no theoretical results in the paper.}
	\item Did you include complete proofs of all theoretical results?
    \answerNA{There are no theoretical results in the paper.}
\end{enumerate}

\item If you ran experiments (e.g., for benchmarks)\dots
\begin{enumerate}
  \item Did you include the code, data, and instructions needed to reproduce the main experimental results (either in the supplemental material or as a URL)?
    \answerNo{We include all all of the data and instructions needed. We will release all code for our benchmarks no later than the publication date.}
  \item Did you specify all the training details (e.g., data splits, hyperparameters, how they were chosen)?
    \answerYes{We provide information about the data splits and hyperparameters in the main paper and supplement. }
	\item Did you report error bars (e.g., with respect to the random seed after running experiments multiple times)?
    \answerNo{Our experiments are too computationally expensive to run multiple times to produce error bars. Our models are meant as baselines that subsequent methods ought to show improvement on.}
	\item Did you include the total amount of compute and the type of resources used (e.g., type of GPUs, internal cluster, or cloud provider)?
    \answerYes{We provide a detailed documentation of the compute cost of each of the steps in the supplement. Some steps, such as the registration of all cameras, are costly. However, believe that by doing them and providing the information to the community, we can collectively save substantial overall compute time across the community.}
\end{enumerate}

\item If you are using existing assets (e.g., code, data, models) or curating/releasing new assets\dots
\begin{enumerate}
  \item If your work uses existing assets, did you cite the creators?
    \answerYes{We build exclusively on the EPIC-KITCHENS 100 dataset. Our use of the dataset is made clear throughout the paper. }
  \item Did you mention the license of the assets?
    \answerYes{When introducing EPIC-KITCHENS, we introduce its license. Our new assets will be released under the same license.}
  \item Did you include any new assets either in the supplemental material or as a URL\@?
    \answerYes{Yes, we provide additional assets that will provide to the reviewers privately and then share publicly upon acceptance.}
  \item Did you discuss whether and how consent was obtained from people whose data you're using/curating?
    \answerYes{When introducing EPIC-KITCHENS, we mention that it was collected with ethics approval and with clear consent from the people who are in the data.}
  \item Did you discuss whether the data you are using/curating contains personally identifiable information or offensive content?
    \answerYes{When introducing EPIC-KITCHENS, we discuss that the data does not contain PII, was reviewed by participants prior to publication, and does not contain offensive content.}
\end{enumerate}

\item If you used crowdsourcing or conducted research with human subjects\dots
\begin{enumerate}
  \item Did you include the full text of instructions given to participants and screenshots, if applicable?
    \answerNA{We did not do any crowdsourced annotation.}
  \item Did you describe any potential participant risks, with links to Institutional Review Board (IRB) approvals, if applicable?
    \answerNA{There are no participant risks for this paper. The dataset we use, however, was collected with ethics board approval.}
  \item Did you include the estimated hourly wage paid to participants and the total amount spent on participant compensation?
    \answerNA{We did not have any paid participants or workers.}
\end{enumerate}

\end{enumerate}

\clearpage
\newpage\appendix
\makeatletter
\renewcommand{\paragraph}{%
  \@startsection{paragraph}{4}%
  {\z@}{-0em}{-0.5em}%
  {\normalfont\normalsize\bfseries}%
}
\makeatother

\etocdepthtag.toc{mtappendix}
\etocsettagdepth{mtappendix}{subsection}
\etocsettagdepth{mtchapter}{none}

\titlespacing*{\section}{0pt}{0.5\baselineskip}{0.3\baselineskip}
\titlespacing*{\subsection}{0pt}{0.5\baselineskip}{0.3\baselineskip}
\titlespacing*{\subsubsection}{0pt}{0.5\baselineskip}{0.3\baselineskip}

In this supplementary material, we first describe the companion video that provides an overview of our dataset (\Cref{sec:sup_video}) and then detail how the data was released (\Cref{sec:sup_data}) along with taking stock of additional information specifically promised in the checklist (\Cref{sec:checklist}). 
Next, we provide additional details on the dataset construction (\Cref{sec:sup_construction}) and on the benchmarks (\Cref{sec:supp_benchmarks}).
We devote a final section (\Cref{sec:supp_ego4d}) to showing that the \dataset pipeline could be applied to reconstructing videos from the Ego4D dataset.

\section{Supplementary video}%
\label{sec:sup_video}

We provide a short video %
in the form of a trailer at \url{https://youtu.be/RcacE26eObE}. It allows to visually assess how challenging the reconstruction problem is and hints at how frame filtering helps. The video also illustrates how the new camera poses complement the existing semantic annotations for this dataset (hands and active objects), showcasing the potential of marrying 3D geometry and video understanding. Additionally, we provide a couple of qualitative results for static and dynamic novel view synthesis, %
one of the benchmark tasks we describe in the paper.

\section{Released data}%
\label{sec:sup_data}

Our dataset is now publicly available with visualisation scripts that enable exploring all the reconstructions and camera poses.

The data can be downloaded from \url{http://epic-kitchens.github.io/epic-fields}.
We released the camera parameters along with sparse point clouds (light-weight version of 10--20MB/video) as well as the full COLMAP database of dense registrations (heavy-weight version). The latter enables comparisons with the dense registrations in \dataset{}, and also allows the use of the COLMAP library and interface for visualisation and exploration.

The webpage also includes links to the visualisation code and to the code to replicate training, inference and evaluation for our benchmarks.

\section{Dataset and benchmark details mentioned in the checklist}%
\label{sec:checklist}

\paragraph{Data splits.} We provide information about the data splits used in the benchmark in \Cref{s:collection}. 

\paragraph{Annotations.} 
We offer two additional sets of manual annotations, on top of those available in VISOR~\cite{Darkhalil22VISOR} to facilitate the assessment of the MG, NeuralDiff, T-NeRF+, and NeRF-W benchmarks.
We employ these annotations as ground truth for evaluation on the UDOS task; nevertheless, we anticipate that they may prove valuable for various applications in future research endeavors.

The first set of annotations serves the dynamic objects.
We provide human body annotations for all evaluation frames, as VISOR exclusively annotates the hands but not the other visible parts of the body.
To achieve this, we identify up to 3 frames per video with visible body parts of the camera wearer.
Using manual points, we employ SAM~\cite{kirillov2023segany} to generate a total of 143 automated human-body annotations.
These frames serve as reference frames for the DeAOT~\cite{yang2022deaot} model pre-trained on YT-VOS to propagate the masks across all evaluation frames. 

The second set of annotations is dedicated to semi-static objects.
VISOR primarily addresses active objects within specific segments of the video, whereas our method aims to evaluate semi-static objects that may have moved at any point during the video. To achieve this, we utilize a fine-tuned MS-DeAOT~\cite{yang2022deaot} on VISOR along with a maximum of 10 VISOR ground truth annotations as reference frames to extend the coverage of semi-static objects across all evaluation frames.
As a result, all objects that have moved during the video are annotated by a mask, on every evaluation frame.

\begin{figure}[t!]
\centering
\includegraphics[width=0.98\textwidth]{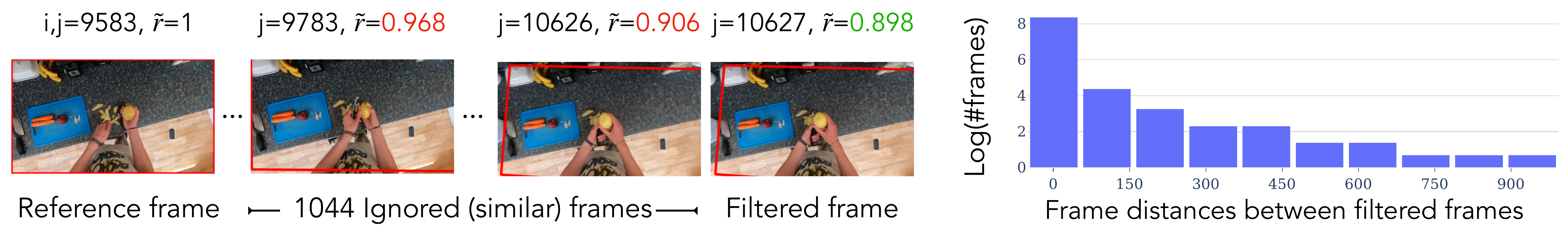}
\caption{\textbf{Filtering frames before reconstruction.} We apply a 2D frame filtering technique to mitigate the oversampling of highly overlapping views (viewpoint distribution skews mentioned in Section 3.2 of the main paper) and to reduce the complexity of the SfM reconstruction. (Left) For a reference frame $i$, we show two of the ignored frames, the next frame after filtering, and their respective overlap $r$ score with the original frame. Filtering discards 1044 frames (ca. 17 seconds) in this case. (Right) Histogram of the distances between frames after filtering (for one video).}\label{fig:frame_filtering}
\vspace{-0.5cm}
\end{figure}

\paragraph{Hyperparameters.} We provide information about the baselines used in our benchmark and their hyperparameters in \Cref{sec:supp_benchmarks}.

\paragraph{Total compute used.} Estimating the precise computational budget of a multi-institution project of this scope is challenging. However, we report the actual computational time specifying the machine used in each case. All resources used were local. The main components of this project were:
\begin{itemize}
    \item \emph{Reconstruction:} As described in \Cref{s:datasetstats}, the reconstruction corresponds to a total of 2264 hours of compute, 1695 hours for the sparse reconstructions and 569 for registration. 
    This was parallelised across two machines with two GPUs each (two 11GB NVIDIA GeForce RTX 2080 Ti for the first machine, 12GB NVIDIA TITAN X and 11GB NVIDIA GeForce GTX 1080 Ti for the second machine).
    \item \emph{NVS, UDOS Benchmarks:} We estimate that running the 3D baselines for D-NVS and UDOS benchmarks required 2400 GPU hours. Experiments on the D-NVS benchmark were carried out using several NVIDIA GPUs on a cluster, including P40, M40, V100, RTX8k and RTX6k. The training required up to 10GB of GPU memory for each experiment. The models for both benchmarks required a total of about 2400 GPU hours. We ran the experiments in parallel on 24 GPUs, resulting in a training time of 4.17 days.  Both D-NVS and UDOS required each 50\% of the total computation.
    \item \emph{MG (UDOS Benchmark)}: We ran this baseline on a single 16GB V100 GPU\@. The total training time is about 5.5 days.
    \item \emph{VOS Benchmark:} The \emph{Fixed in 3D} baseline requires next-to-no compute --- homography fitting on SIFT features is calculated during the reconstruction step. However, training STM and XMEM took 1.2 and 1.4 days respectively on a single 16GB V100 GPU\@.
\end{itemize}
We expect that by providing both sparse and dense reconstructions to the whole community, this effort will greatly reduce computation time for all the dataset users.

\section{Additional details on the dataset construction}%
\label{sec:sup_construction}

\subsection{Frame filtering}
As discussed in Section 3.2 in the main paper, we downsampled videos to reduce the viewpoint skew that is common for ego-centric videos. The filtering discards on average 81.8\% of all frames and allows the SfM pipeline to focus on more diverse views. \Cref{fig:frame_filtering} visualises the filtering process using an example. The shown frame range contains many views that are similar to each other. The filtering discards 1044 redundant frames between frames $j = 9583$ and $j=10627$. 
The figure also shows a histogram of distances between filtered frames.

\subsection{Dataset statistics}%
\label{s:datasetstats}

\begin{figure*}[t]
\centering
\includegraphics[width=\textwidth]{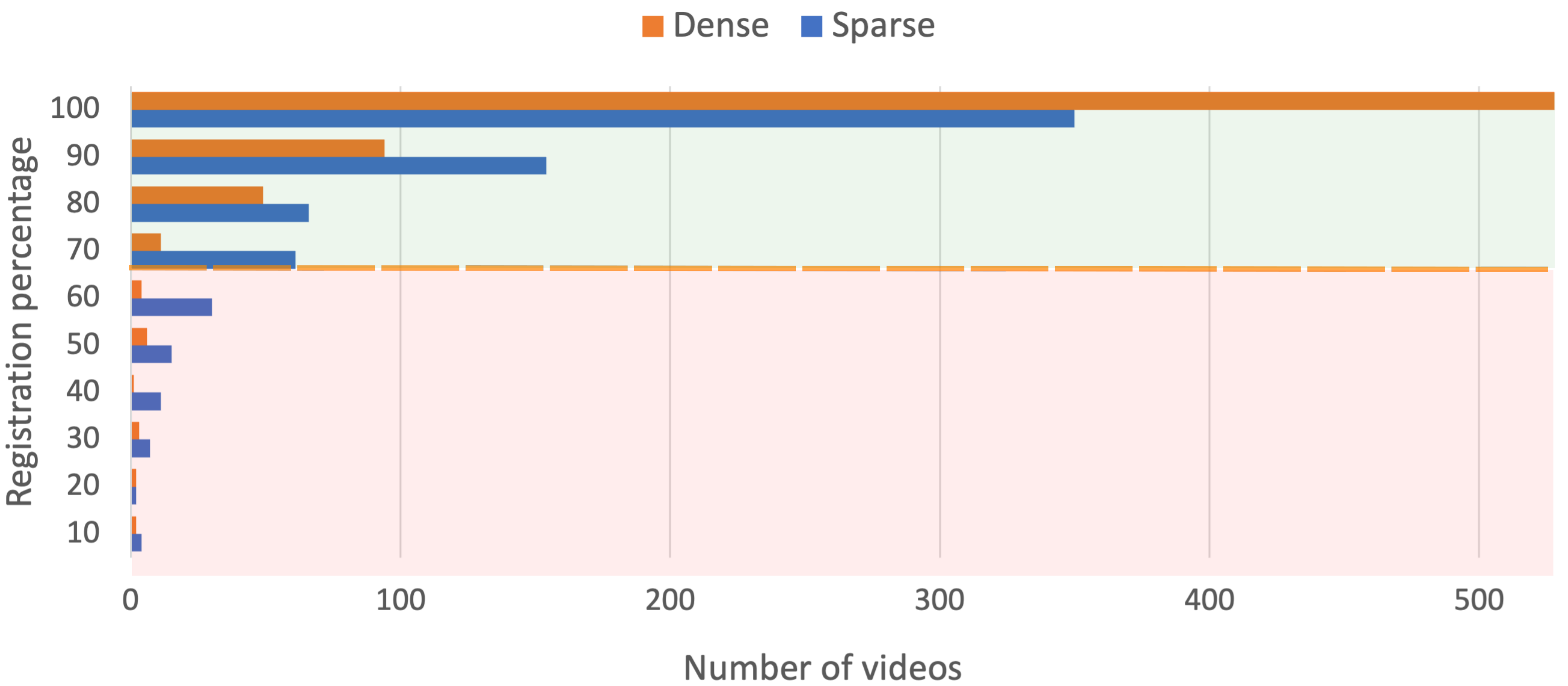}
\caption{\textbf{Percentages of registered frames.} The dashed line specifies the threshold of the minimum dense registration rate to accept the reconstruction, otherwise, it would be considered a failure.}\label{fig:registration_histogram}
\end{figure*}

\begin{figure*}[t]
\centering
\includegraphics[width=0.7\textwidth]{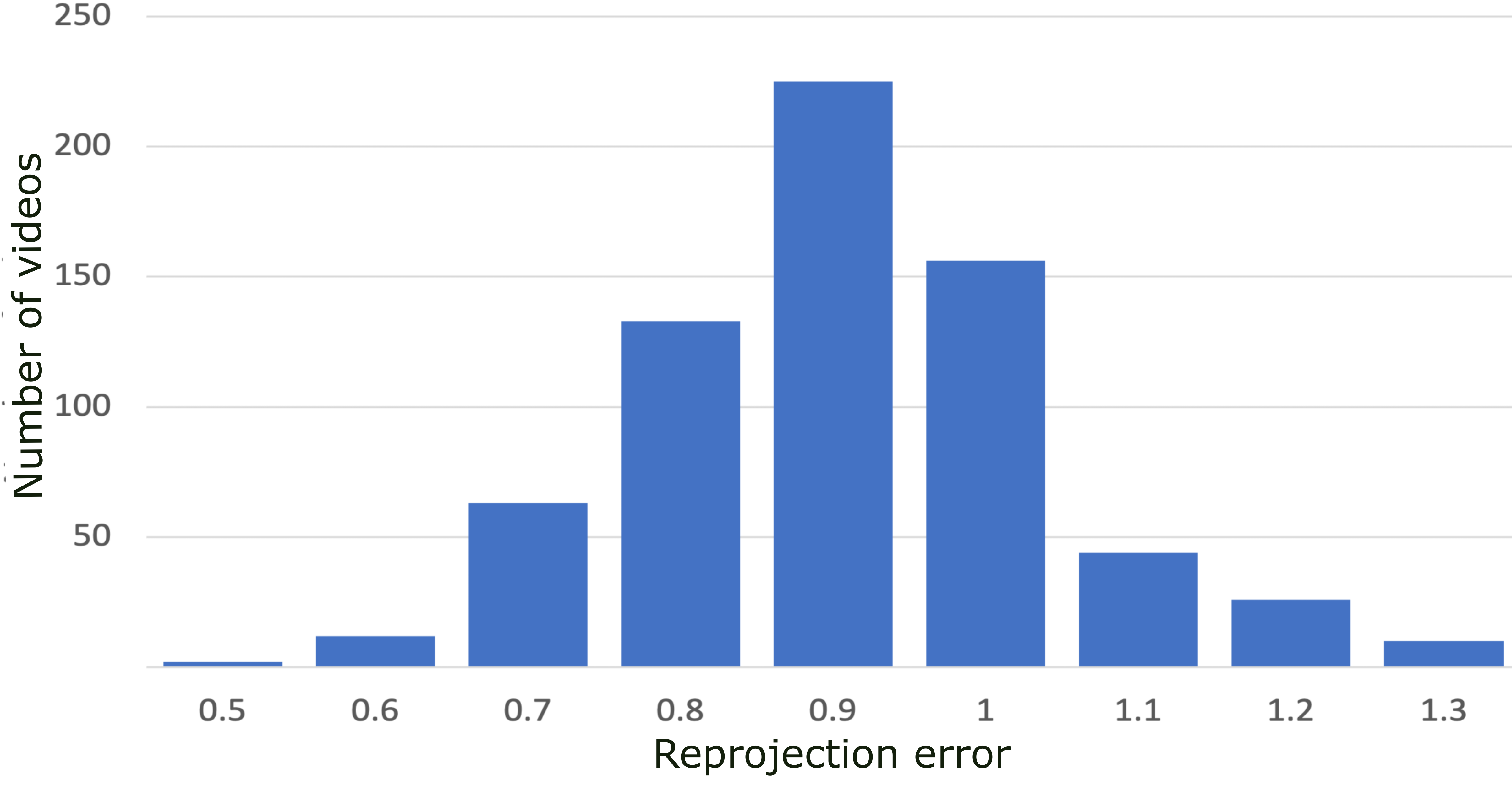}
\caption{\textbf{Average reprojection error of \dataset{}.} The majority of our reconstructions have an average reprojection error lower than 1.}\label{fig:reproj_error}
\end{figure*}

\begin{figure*}[t]
\centering
\includegraphics[width=0.8\textwidth]{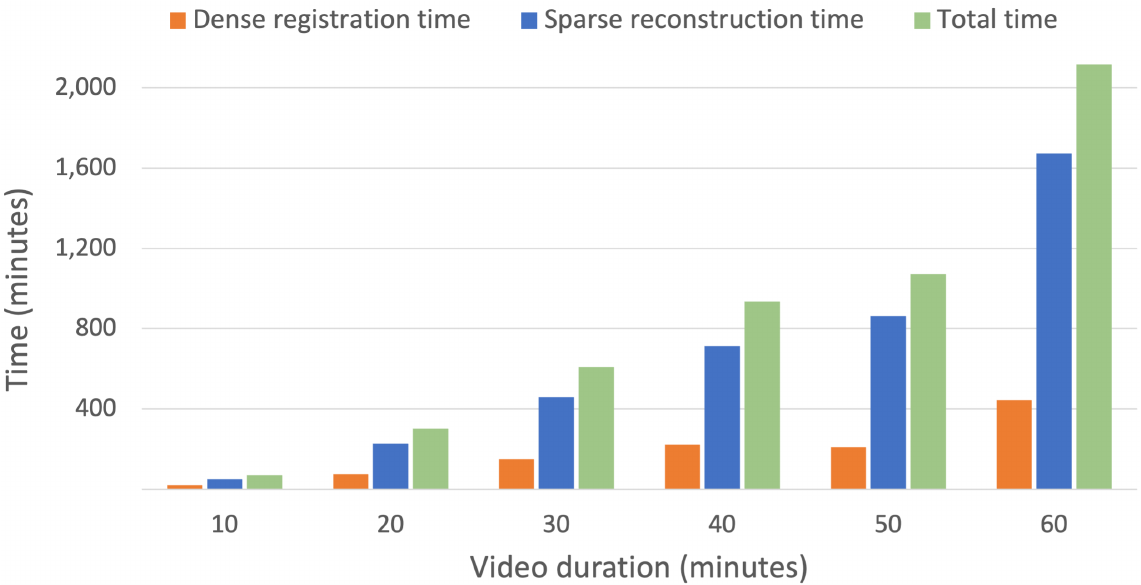}
\caption{\textbf{Reconstruction time per video length.} We plot time for the sparse reconstruction (blue), registration time to obtain the dense camera poses (orange) and total reconstruction time (green) for different video durations. While the time for registration is almost linear, the reconstruction time increases non-linearly as a function of the video length, mainly because of bundle adjustment.} %
\label{fig:reconstruction_time}
\end{figure*}

\begin{figure*}[t]
\centering
\includegraphics[width=\textwidth]{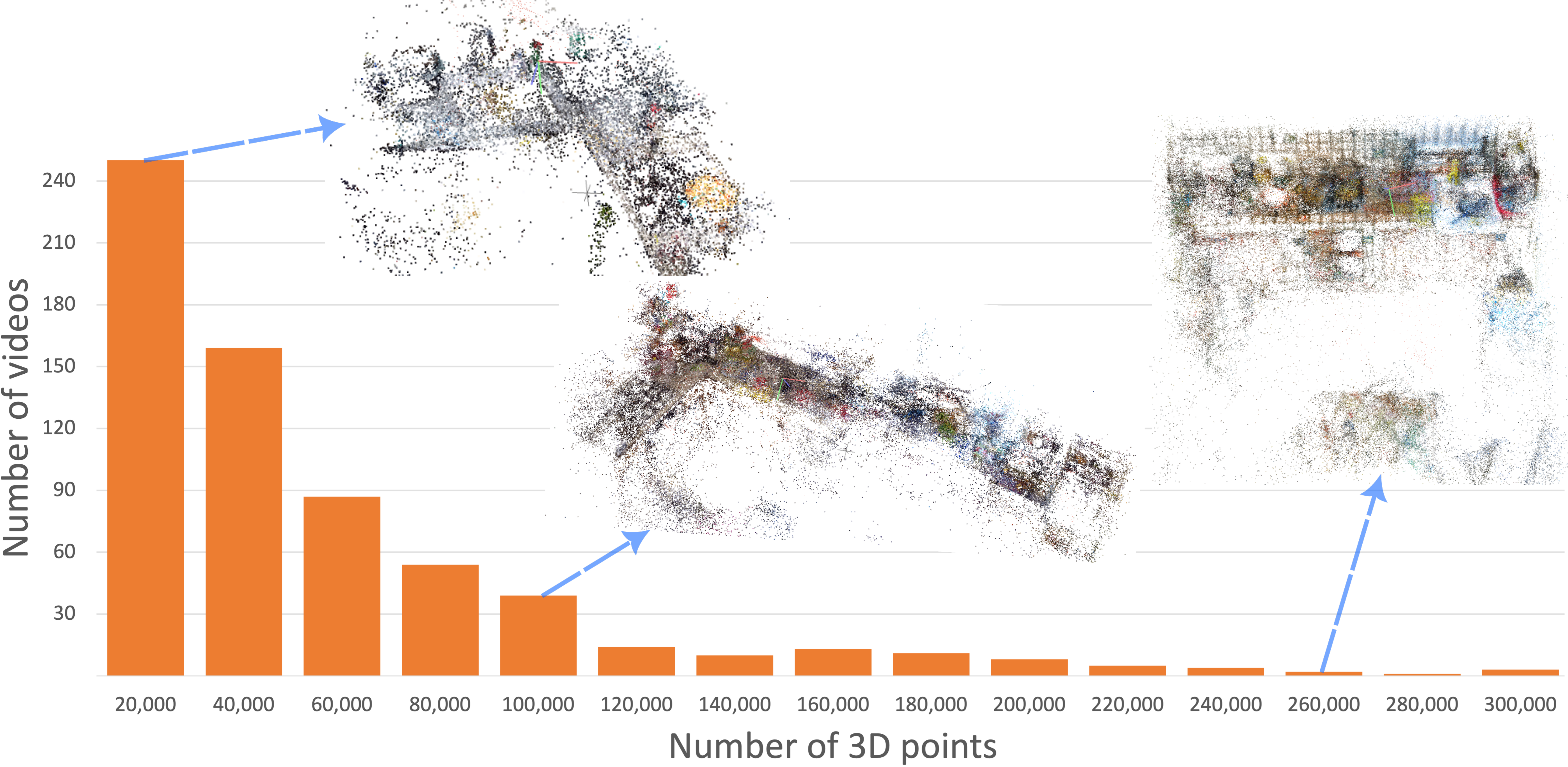}
\caption{\textbf{Number of 3D points histogram.} The majority of our reconstructions generate fewer than 40,000 points that are enough to represent the kitchen. However, some reconstructions have more than 100,000 points, we include the point clouds for each points range showing the fine details covered by having more points.}\label{fig:points_histogram}
\end{figure*}

\begin{figure*}[t]

    \centering
    \begin{subfigure}[b]{0.32\textwidth}
    \centering
        \includegraphics[width=\textwidth]{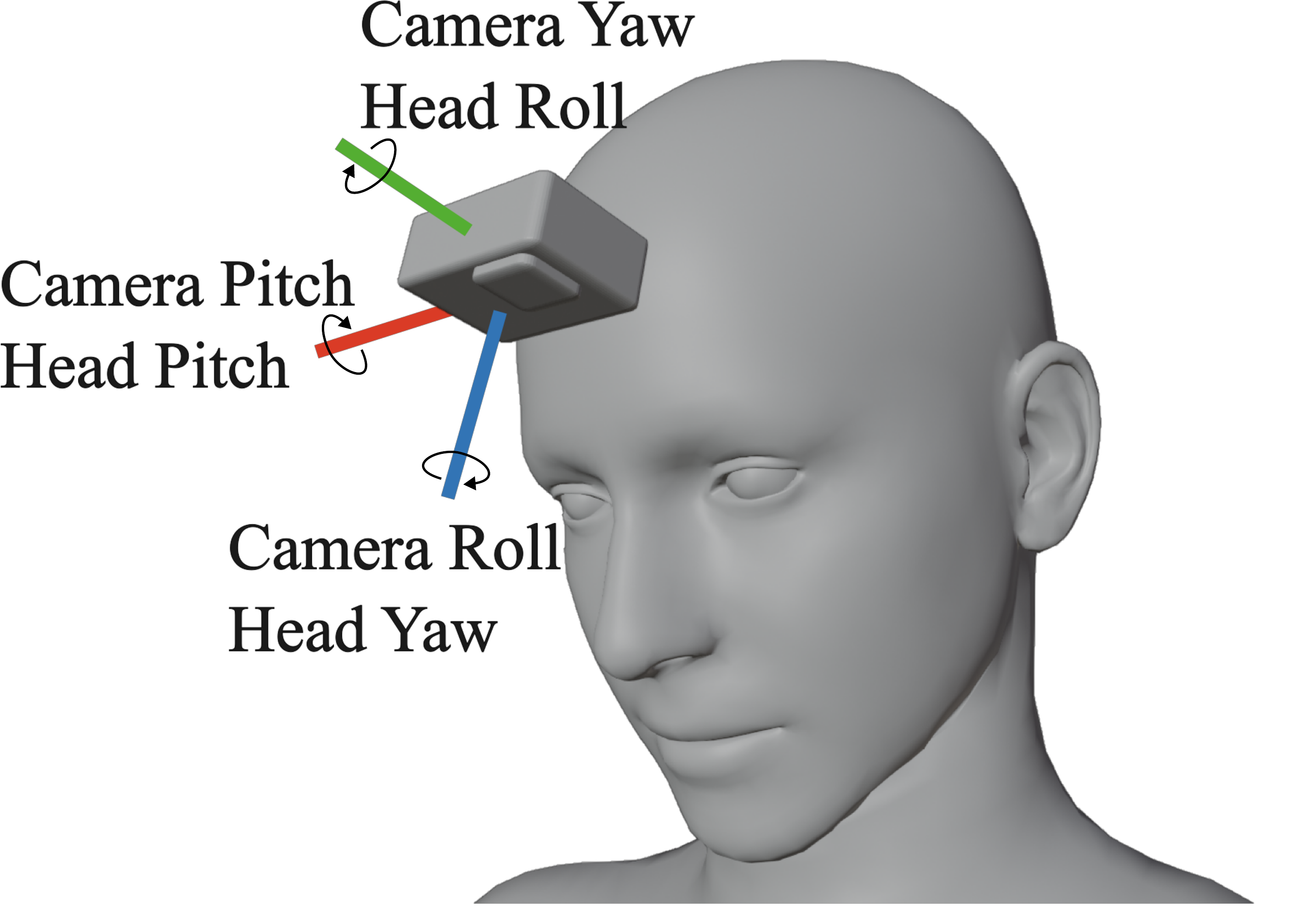}
        \caption{The correspondence of \\
        Yaw/Pitch/Roll between the \\
        camera and the head.}
     \end{subfigure}
     \hfill
     \centering
     \begin{subfigure}[b]{0.22\textwidth}
         \centering
         \includegraphics[width=\textwidth]{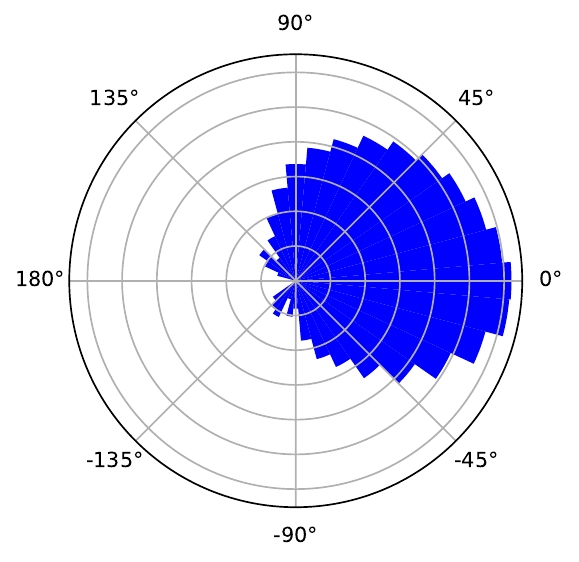}
         \caption{Camera Pitch / Head \\
         Pitch / Head Raising \\
         Up-or-Down}
     \end{subfigure}
     \hfill
     \begin{subfigure}[b]{0.22\textwidth}
         \centering
         \includegraphics[width=\textwidth]{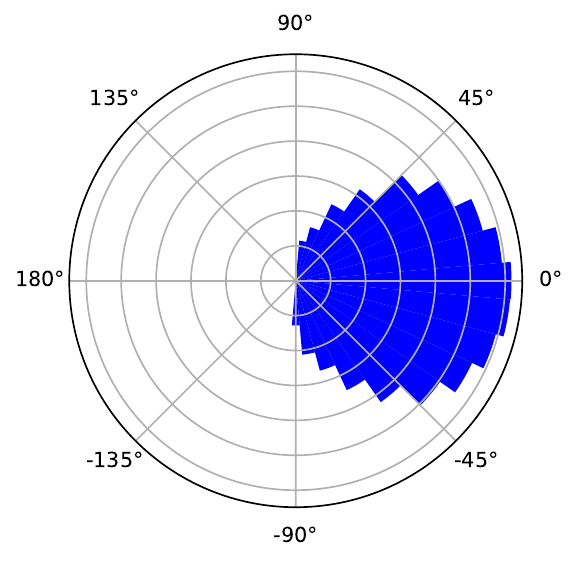}
         \caption{Camera Yaw / Head \\
         Roll / Head Tilting \\
         Left-or-Right}
     \end{subfigure}
     \hfill
     \begin{subfigure}[b]{0.22\textwidth}
         \centering
         \includegraphics[width=\textwidth]{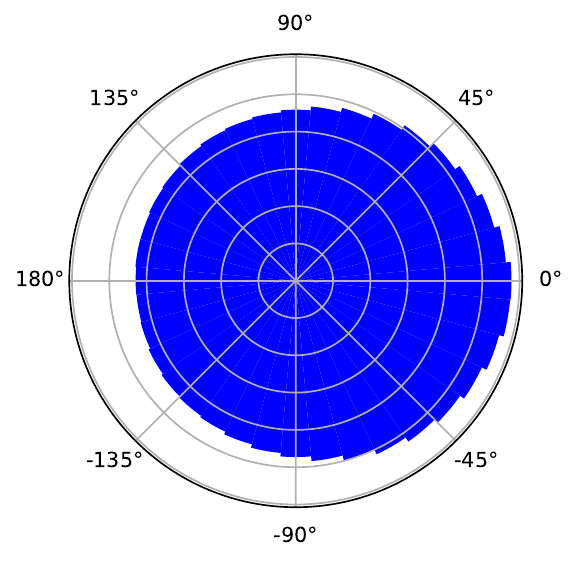}
         \caption{Camera Roll / Head \\
         Yaw / Neck Turning \\
         }
     \end{subfigure}
\caption{Camera mounting arrangement (a) and the log-scale polar histogram of the three camera orientation parameters in the dataset (b-d).}\label{fig:camera_angles}
\end{figure*}

\textbf{How do we accept/reject a reconstruction?} After producing the sparse reconstructions, we register all the frames; we then consider the videos with at least 70\% dense registration rate. The histogram for both sparse and dense reconstructions is depicted in \Cref{fig:registration_histogram}. The majority of our reconstructions exhibit a dense registration rate exceeding 80\%. In total, we successfully reconstructed 671 out of the 700 \epic videos, with average registration rates of 84.1\% and 92.0\% for the sparse and the dense reconstructions respectively.
This is because we specifically select frames during transitions between kitchen hotspots for accurate reconstruction. This explains the higher registration rate for dense reconstructions.

\textbf{Metrics for reconstruction quality.} We use the common SfM metrics to assess the quality of the reconstructions.
\Cref{fig:reproj_error} shows the histogram of the reprojection error of all the reconstructions. The average and maximum reprojection errors are 0.87px and 1.3px respectively. We use an image resolution of 456$\times$256 to obtain the reconstructions and to calculate the reprojection errors.

\textbf{How long does the reconstruction pipeline take?} In \Cref{fig:reconstruction_time}, for different video durations, we report the time required for the sparse reconstruction, for registration to obtain the dense reconstruction, and the total reconstruction time. As the length of the videos increases, the sparse reconstruction time follows a non-linear growth pattern. Overall, the sparse reconstruction and the registration processes took 1695 and 569 computation hours, respectively. We parallelise the pipeline on 2 machines with 2 GPUs each.

\textbf{How large are these reconstructions?} 
\Cref{fig:points_histogram} displays a histogram representing the number of 3D points in the sparse reconstructions, along with three example point clouds derived from reconstructions with varying numbers of points. These demonstrate the complexity of our reconstructions, which are capable of covering entire kitchens with fine-grained details. On average, each reconstruction consists of around 45,000 3D points.

\textbf{Reasons for the reconstruction failures.} While our reconstruction failure rate is only 4\%, we examined the primary causes of these failures. These are mainly attributed to very short videos with large scene coverage, and challenging lighting conditions.
(1) In the case of very short videos with large scene coverage, \eg a person just walking through the kitchen to retrieve one item and then walking out again, COLMAP often encounters difficulties %
due to the insufficient quantity of features and viewpoints. The median duration for the unsuccessful reconstructions is 1.5 minutes, compared to 6 minutes for the successful ones. This problem is exacerbated when the brief video captures a multitude of different locations within the kitchen, switching rapidly between these. (2) %
A couple of failure cases were linked to videos recorded under very low lighting, which led to a %
poor %
quality set of features to match. The average number of observed features per image for these unsuccessful videos was 198, compared to an average of 358 features per image for successful reconstructions.

\textbf{Distribution of camera orientations.} \Cref{fig:camera_angles} displays histograms representing the distribution of \textit{relative} camera orientations of all \dataset frames. Each frame uses the mean camera orientation within the video as reference. The histograms reveal that \dataset contains diverse camera motions that are a result of natural head motions, such as looking up/down or tilting left/right.
It is important to note the distinction between the camera orientations due to the particular camera mounting in EPIC-KITCHENS, illustrated in the figure. We thus particularly note camera motions and how they correspond to head motion given the specified mounting.

In summary, the figure shows larger head motion looking up (compared to the average camera orientation) than looking down, a balanced tilting as well as full 360 coverage of the kitchen by the body and/or head rotating in the scene.

\subsection{Statistics of the benchmark splits}%
\label{s:collection}

We provide statistics of the splits for the D-NVS task of our benchmark in \Cref{tab:dataset_split_stats}. In UDOS, the objective is to segment dynamic and semi-static objects in videos without relying on supervision from ground-truth segmentations during training. Thus, \emph{all} frames are observed during training. Evaluation frames are the same as for the D-NVS task. For the VOS task, we use the train/val splits published as part of the VISOR VOS benchmark (See~\cite{Darkhalil22VISOR} Sec.~5.1). 

For D-NVS, we divide the evaluation frames for each video equally between the validation and test sets, taking every other frame from both \textit{In-Action} and \textit{Out-of-Action} frames. 
Each video contains evaluation frames spanning all difficulty tiers (easy, medium, hard).
The size of the validation and test sets corresponds to only a fraction of the number of training frames due to strict constraints on the sampling of evaluation frames, which include high variability in viewpoints and a minimum time gap between the training and test/validation frames as described in Section 4.1 of the main paper.

For the \textit{Hard (In-Action)} and \textit{Medium (Out-of-Action)} settings, this time gap is set to 1s, which introduces increased difficulty for rendering novel views, %
since a significant portion of an activity might have taken place and neural rendering approaches would have to interpolate motion to account for this.
While this is indeed a challenging task, it provides a unique opportunity for further explorations in neural rendering. 
We can account for the ambiguity that this choice introduces in two ways: resort to an evaluation protocol that accounts for that (e.g., best-of-K prediction) or accept that pixel predictions will have to be approximate for dynamic pixels and still measure the PSNR score. While the latter is not perfect, it is still reasonable for most 1s gaps and is much simpler than alternatives. The preference for this choice is also common in other ambiguous prediction tasks; for example, in 
the GTA-IM benchmark, where 3D path error is estimated after 0.5, 1, 1.5, and 2s~\cite{cao2020long},
the TrajNet benchmark, where prediction is estimated for 4.8s from the observed frame~\cite{becker2018evaluation},
and the future hand prediction task in Ego4D, which uses a time gap of 1.5s from the observed frame~\cite{grauman2022ego4d}.

For the \textit{Easy (Out-of-Action)} setting, there is no temporal gap between training and evaluation frames and no specific action taking place.
Consequently, both the complexity for rendering novel views and the ambiguity in evaluation are reduced for this subset of frames.
This simplified setup parallels existing NVS benchmarks.

\section{Additional training details for benchmarks}%
\label{sec:supp_benchmarks}

We now provide precise hyperparameters for the baselines used in the NVS and UDOS benchmarks.
We provide full code for reproducing these results with the publication.

\paragraph{NeRF-W, T-NeRF+, NeuralDiff.} We base our implementation of all 3D baselines on the codebase from NeuralDiff~\cite{tschernezki21neuraldiff} and merge the other two approaches into the same PyTorch~\cite{pytorch} codebase to align all training and evaluation details between models. We use the same training setup as in NeuralDiff, which involves training one model per baseline on each scene, taking approximately 12 hours using one NVIDIA Tesla P40 per experiment. Furthermore, the models are trained with hierarchical sampling (with a coarse and fine model as in the typical NeRF setting) and with a batch size of 1024. We train with the Adam optimizer for 10 epochs and an initial learning rate of $5 \times 10^{-4}$ that is adjusted during the training with a cosine annealing schedule.

\paragraph{MG.} We use the provided code and train the model on our training split frames, jointly, for 135k iterations with a batch size of 32 and a learning rate of $5 \times 10^{-4}$.

\paragraph{STM and XMEM.} For STM, we finetune a pretrained COCO~\cite{lin2014microsoft} model on VISOR for 400K iterations with a batch size of 32 and a learning rate of $1 \times 10^{-5}$. For XMEM, we use the pretrained YoutubeVOS~\cite{vos2018} model published in the XMEM paper and finetune it on VISOR for 100K iterations, with a batch size of 16 and a decaying learning rate initialised with $1 \times 10^{-5}$.

\begin{table*}[t]
\caption{\textbf{\dataset splits statistics}. We summarise the frame count and average frames per video for each split and for different difficulties (Easy, Medium, Hard). The number of frames for the validation and test sets is only a fraction of the training frames. This is due to strict constraints on the sampling of evaluation frames such as a high variety of viewpoints and the minimum time frame between train and test/validation frames. The train frames are fixed, regardless of the difficulty level.}%
\label{tab:dataset_split_stats}
\resizebox{1.0\linewidth}{!}{
\begin{tabular}{ccccccccc}
\toprule
 & \multicolumn{2}{c}{In-Action (Easy)} &  \multicolumn{2}{c}{Out-of-Action (Medium)} & \multicolumn{2}{c}{Out-of-Action (Hard)} &\multicolumn{2}{c}{Total} \\ 
\cmidrule(lr){2-3} \cmidrule(lr){4-5} \cmidrule(lr){6-7}\cmidrule(lr){8-9}
 &\multicolumn{1}{c}{\#frames}&\multicolumn{1}{c}{average} &\multicolumn{1}{c}{\#frames}&\multicolumn{1}{c}{average}&\multicolumn{1}{c}{\#frames}&\multicolumn{1}{c}{average}&\multicolumn{1}{c}{\#frames}&\multicolumn{1}{c}{average} \\ 
\midrule

 Train & \cellcolor[gray]{0.8} --- & \cellcolor[gray]{0.8} ---& \cellcolor[gray]{0.8} ---& \cellcolor[gray]{0.8} ---& \cellcolor[gray]{0.8} ---& \cellcolor[gray]{0.8} --- & 103,571 & 2071.42\\ 
Val & 3,448 & 68.96 & 657 & 13.14 & 305 & 6.1 & 4,410 & 88.2 \\
Test & 3,461 & 69.22 & 695 & 13.9 & 289 & 5.78 & 4,445 & 88.9 \\
\bottomrule
\end{tabular}}
\end{table*}

\begin{figure*}[t]
\centering
\includegraphics[width=\textwidth]{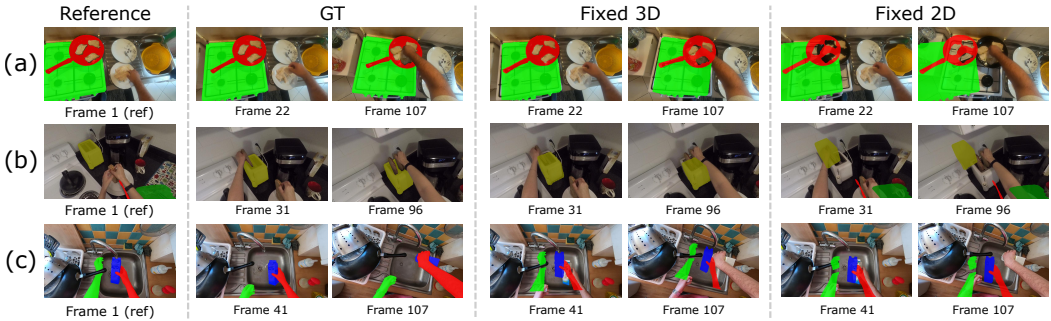}
\caption{\textbf{Qualitative results for Semi-Supervised VOS.} We show samples with multiple objects. In scenarios (a) and (b), the \textit{Fixed 3D} baseline effectively handles static objects, whereas the \textit{Fixed 2D} falters due to camera movement. Conversely, in scenario (c), both strategies prove unsuccessful as the objects are in motion, invalidating the presumption of fixed objects in either 3D (\textit{Fixed 3D}) or 2D (\textit{Fixed 2D}), hence their failure.}\label{fig:vos_qualititve_supp}
\end{figure*}

In the main paper, we include some qualitative results for the VOS challenge for a single object. We add more examples showing multi-object segmentation in \Cref{fig:vos_qualititve_supp}. The figure shows samples of failures of \textit{Fixed 2D} in scenarios (a) and (b) and a case when both \textit{Fixed 2D} and \textit{Fixed 3D} fail to segment the dynamic objects (c).

\section{\dataset pipeline for Ego4D videos}%
\label{sec:supp_ego4d}

\begin{figure*}[t]
\centering
\includegraphics[width=\textwidth]{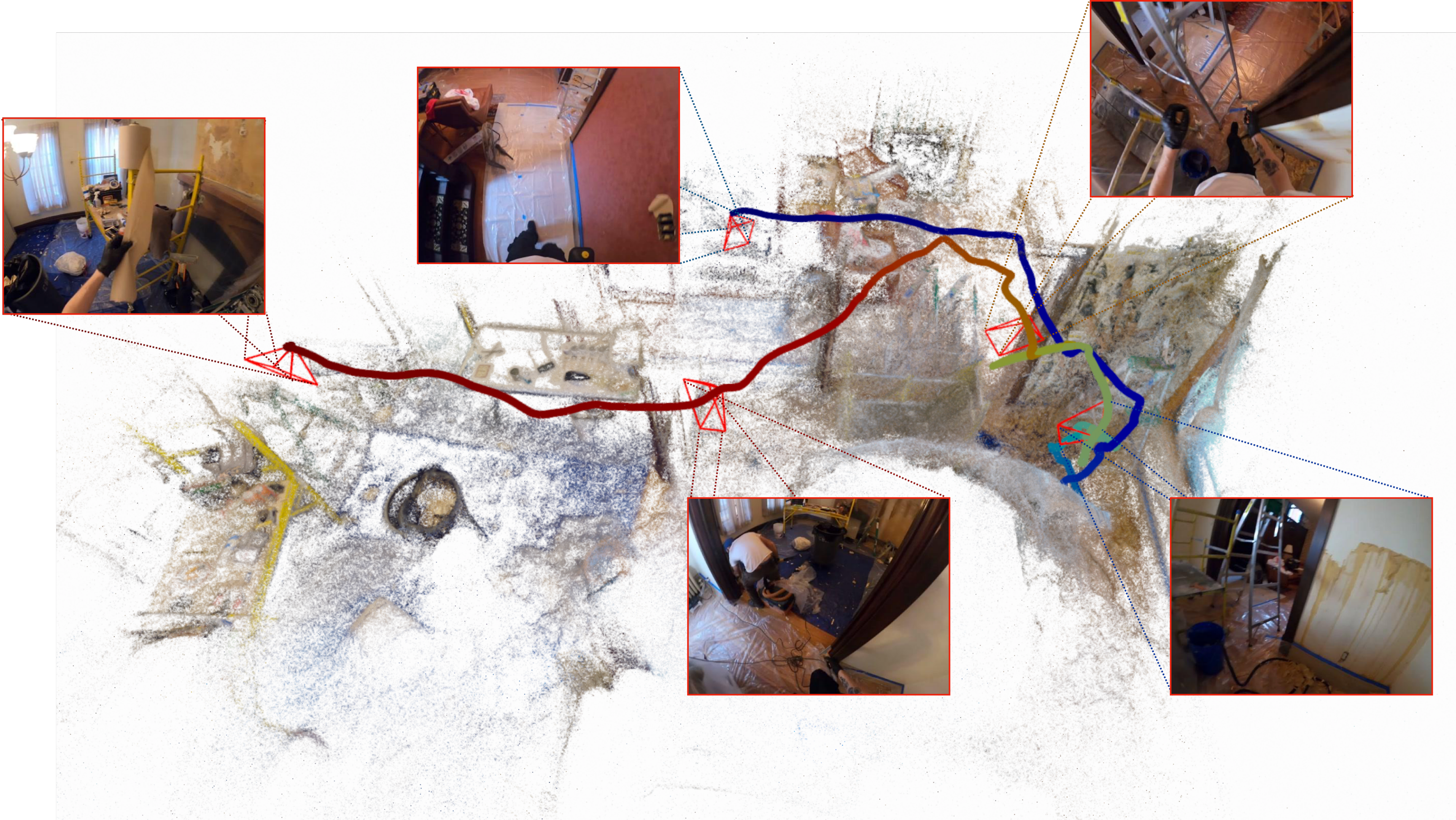}
\caption{Visualisation of the 3D reconstruction for one video of the Ego4D dataset capturing building and refurbishment activities, with camera estimated using the \dataset pipeline}\label{fig:ego4d}
\end{figure*}

While our reconstruction pipeline addresses several difficulties that are inherent to the videos of EPIC-KITCHENS~\cite{Damen2022RESCALING}, we can also apply it to other ego-centric videos such as the ones from Ego4D~\cite{grauman21ego4d}.
Using the pipeline as is, we can estimate camera poses for Ego4D videos that are about cooking and construction/building.
We showcase this through an example in \Cref{fig:ego4d} and two videos of reconstructions and camera tracks:

\begin{itemize}
\item Task: Construction –-- 35 minutes of decorating and refurbishment. The video at \url{https://youtu.be/EZlayZIwNgQ} contains situations of challenging camera pose estimation including the camera wearer on a ladder (01:29, 05:07), kneeling down (16:14), as well as drinking and navigating the scene (27:25) amongst many interesting poses. (Ego4D video a2dd8a8f-835f-4068-be78{-}99d38ad99625, source: CMU US)
\item Task: Cooking –-- 10 minutes. The corresponding video can be found at \url{https://youtu.be/GfBsLnZoFGs} (Ego4D video 18f5c2be-cb79{-}46fa-8ff1-e03b7e26c986).
\end{itemize}

\end{document}